\documentclass{article} 
\usepackage[table]{xcolor} 
\usepackage{iclr2026_conference,times}

\usepackage{amsmath,amsfonts,bm}

\def\eqref#1{equation~\ref{#1}}

\def\1{\bm{1}}

\DeclareMathAlphabet{\mathsfit}{\encodingdefault}{\sfdefault}{m}{sl}
\SetMathAlphabet{\mathsfit}{bold}{\encodingdefault}{\sfdefault}{bx}{n}

\usepackage{url}
\usepackage{arydshln}    
\usepackage{booktabs}       
\usepackage{multirow}       
\usepackage{adjustbox}
\usepackage{amssymb}
\usepackage{makecell}
\usepackage{caption}
\usepackage{tabularx}

\usepackage{soul}           
\usepackage[switch]{lineno} 
\usepackage{graphicx}       
\usepackage{wrapfig}
\usepackage{subcaption}

\usepackage{algorithm}
\usepackage{algorithmicx}
\usepackage{algpseudocode}

\title{Mitigating Visual Hallucinations via Semantic Curriculum Preference Optimization in MLLMs}

\author{
Yuanshuai Li\textsuperscript{1}\hspace{0.1em}\thanks{Equal contribution.} \quad
Yuping Yan\textsuperscript{1}\hspace{0.1em}\footnotemark[1] \quad
Junfeng Tang\textsuperscript{1} \quad
Yunxuan Li\textsuperscript{2} \quad
Zeqi Zheng\textsuperscript{1} \quad
Yaochu Jin\textsuperscript{1}\hspace{0.1em}\thanks{Corresponding author.} \\
\textsuperscript{1}School of Engineering, Westlake University, Hangzhou, China \\
\textsuperscript{2}School of Cyberspace Security, Nanjing University of Science and Technology, Nanjing, China \\
\texttt{liyanshuai@westlake.edu.cn}
}

\iclrfinalcopy 
\begin{document}

\maketitle
\pagestyle{plain}   
\thispagestyle{plain}

\begin{abstract}
Multimodal Large Language Models (MLLMs) have significantly improved the performance of various tasks, but continue to suffer from visual hallucinations, a critical issue where generated responses contradict visual evidence. While \textit{Direct Preference Optimization} (DPO) is widely used for alignment, its application to MLLMs often fails to capture fine-grained semantic differences and encourages shortcut learning. To address these challenges, we propose \textit{Semantic Curriculum Preference Optimization} (SCPO), a novel framework for MLLM alignment. SCPO employs a progressive, easy-to-hard curriculum built upon our \textit{Semantic Curriculum Preference Pairs} dataset, which provides fine-grained semantic contrasts sorted by difficulty. This curriculum is trained with a dynamic reference model and a novel symmetric, bidirectional objective to facilitate simultaneous learning from both textual and visual preferences. To our knowledge, SCPO is the first framework to unify semantics, symmetry, and curriculum for MLLMs alignment, effectively mitigating visual hallucinations. Extensive experiments on LLaVA models across various scales and versions validate that SCPO demonstrates superior performance compared to baseline models on multiple hallucination benchmarks, reducing the hallucination rate by up to 62.9\%. Moreover, evaluations on generalized benchmarks show that SCPO improves factuality while preserving general capabilities, with its performance remaining stable across general vision-language benchmarks.
\end{abstract}

\section{Introduction}

Multimodal large language models (MLLMs) \citep{liu2024llavanext,gpt4V,liu2024visual} have demonstrated remarkable capabilities in tasks requiring fine-grained perception, such as visual question answering and image captioning \citep{yuAligningMultimodalLLM2025}. Despite these advances, their reliability is compromised by the persistent challenge of visual hallucination. During inference, MLLMs often rely too heavily on the semantic priors embedded in the language model, while failing to faithfully ground these priors in fine-grained visual evidence from the input image \citep{li2022blip,kalai2025languagemodelshallucinate,xu2024hallucination}. This misalignment between language and vision produces outputs that contradict the actual visual content, thereby seriously limiting their applicability in high-stakes scenarios such as autonomous driving and medical image diagnosis \citep{yu2024rlhf}.

\begin{figure}[htp]
\centering
\includegraphics[width=\linewidth]{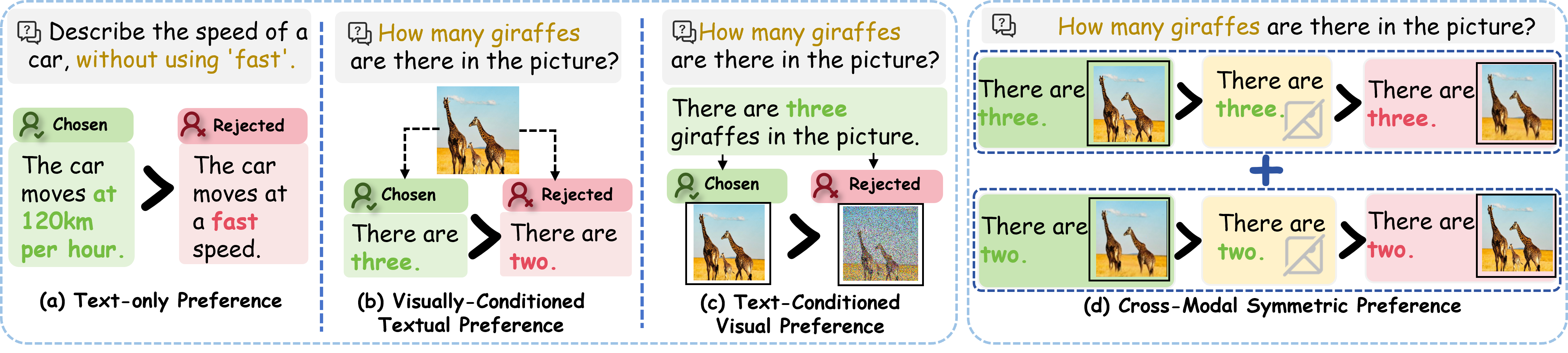}
\vspace{-5mm}  
\caption{
    Comparison of different DPO-based alignment strategies.
    \textit{(a) Text-only Preference}: preference is defined over textual responses.
    \textit{(b) Visually-Conditioned Textual Preference}: preference is defined over textual responses for a fixed image.
    \textit{(c) Text-Conditioned Visual Preference}: preference is defined over images for a fixed response.
    \textit{(d) Cross-Modal Symmetric Preference}: preference is defined symmetrically over complete image-text pairs.
}
\label{fig:fig1}
\vspace{-6.1mm}
\end{figure}

To address this challenge, a common approach is to align models with human preferences using frameworks like Reinforcement Learning from Human Feedback (RLHF)~\citep{christiano2017deep}. RLHF leverages human preferences to guide model optimization, typically by first collecting preference pairs and then employing reinforcement learning techniques such as Proximal Policy Optimization (PPO) to improve output quality and reliability \citep{schulman2017proximal}. Building on this, \textit{Direct Preference Optimization (DPO)} \citep{rafailov2023direct} simplifies the pipeline by skipping the reward modeling and reinforcement learning stages, directly training on \emph{``chosen--rejected''} preference pairs (Figure~\ref{fig:fig1}(a)). Its effectiveness in enhancing instruction-following and reducing undesirable outputs in language models~\citep{touvron2023llama} has made it a leading paradigm for preference-based alignment.

Motivated by DPO's success in textual scenarios, recent work has extended it to MLLMs for finer-grained vision-language alignment. Initial efforts largely mirrored the unimodal paradigm, constructing preference pairs from textual responses, while the visual input remained fixed~\citep{zhao2023beyond,yu2024rlhf,yu2024rlaif,yang2025mitigating}, as illustrated in Figure~\ref{fig:fig1}(b). In this setup, models primarily learn to distinguish factual descriptions from hallucinated ones, but visual evidence is not explicitly incorporated into preference judgment. To overcome this limitation, subsequent work introduced pairs of visual preferences by keeping the textual instruction and \emph{``chosen''} response constant while generating a \emph{``rejected''} image through perturbations of the original~\citep{zhou2024aligning}. This design directly contrasts the original and perturbed images, allowing visual cues to be incorporated more explicitly into preference optimization (Figure~\ref{fig:fig1}(c)). Although incorporating visual preferences is a step forward, current methods are still constrained by several key limitations as discussed below.

First, the current negative sample construction lacks semantic sharpness. Existing methods typically generate ``rejected'' images through random perturbations such as noise injection or cropping \citep{fu2025chip,wang2024mdpo,jiang2024hallucination}. These operations mainly distort low-level pixels while leaving high-level semantics intact, causing models to learn trivial cues of image ``damage'' rather than developing sensitivity to fine-grained cross-modal distinctions.

Second, existing optimization objectives inherently bias alignment toward the image modality, reducing it to a one-sided preference. As shown in prior work \citep{zhou2024aligning,yang2025mitigating,fu2025chip}, existing methods tend to bias learning toward the original image while neglecting symmetric consistency across vision and language. This asymmetry not only undermines true bidirectional grounding but also fosters the learning of spurious correlations, ultimately resulting in brittle and unreliable alignment.

Finally, existing alignment methods typically employ a static single-pass training paradigm. They generally treat the preference dataset as a single collection of data, training the model on a random mixture of samples with varying levels of difficulty. This approach overlooks the benefits of a structured learning process, which can lead to inefficient training and suboptimal final performance~\citep{gao2025principled}. Moreover, by keeping the reference model fixed throughout training, these methods can suffer from significant distribution changes as the policy model improves, potentially limiting the full extent of alignment~\citep{yuan2024self,xu2024dpo,xiongiterative}.

To address these challenges, we propose \textit{Semantic Curriculum Preference Optimization (SCPO)}, a novel alignment framework designed to improve vision-language grounding systematically. SCPO integrates semantically challenging preference pairs, a symmetric and bidirectional optimization objective, and a dynamic curriculum-based training strategy. This unified approach enables robust hallucination mitigation while preserving general model capabilities. Our main contributions are as follows.

\vspace{-3mm}
\begin{enumerate}
\itemsep -3pt
    \item We construct the \textit{Semantic Curriculum Preference Pairs (SCPP)}, a large-scale dataset for fine-grained vision-language alignment. SCPP is built via a novel pipeline that quantifies sample difficulty, enabling an easy-to-hard curriculum to build from foundational grounding to complex semantic reasoning.

    \item We propose a novel \textit{symmetric and bidirectional objective} for preference optimization. Our objective first learns from complementary textual (Figure~\ref{fig:fig1}(b)) and visual (Figure~\ref{fig:fig1}(c)) preferences, and then enforces robust grounding through a symmetric framework (Figure~\ref{fig:fig1}(d)). This unified approach prevents the model from relying on shortcut learning and encourages a deeper semantic understanding.

    \item We introduce the \textit{SCPO framework}, which systematically integrates a difficulty-based curriculum with dynamic reference model updates for MLLM alignment. Extensive experiments demonstrate that SCPO significantly reduces hallucinations and establishes new state-of-the-art performance on multiple hallucination benchmarks.
\end{enumerate}
\vspace{-5mm}

\section{Preliminaries}

Direct Preference Optimization (DPO) \citep{rafailov2023direct} is a reinforcement-learning-free approach to aligning language models with human preferences. It is derived from the KL-regularized reward maximization objective commonly used in RLHF:

\vspace{-10pt}
\begin{equation}
    \max_{\pi} \; \mathbb{E}_{x \sim \mathcal{D}, y \sim \pi(y|x)} [r(x, y)] 
    - \beta D_{\mathrm{KL}}\big(\pi(y|x) \,\|\, \pi_{\text{ref}}(y|x)\big),
    \label{eq:rlhf_objective}
\end{equation}
\vspace{-10pt}

where $\pi$ denotes the policy to be optimized, $\pi_{\text{ref}}$ is a fixed reference policy, $r(x,y)$ is a reward function, $\beta > 0$ is a regularization coefficient that controls the KL penalty, and $\mathcal{D}$ is a dataset of training prompts $x$. This objective allows for aligning a model with human preferences without requiring explicit reward modeling or reinforcement learning.

The optimal policy $\pi^*$ induces a closed-form mapping from the policy to the reward:

\begin{equation}
    r(x, y) = \beta \log \frac{\pi^*(y|x)}{\pi_{\text{ref}}(y|x)} + \beta \log Z(x),
    \label{eq:reward_mapping}
\end{equation}

where $Z(x)$ is a partition function that depends only on $x$.

Human preferences are typically observed through pairwise comparisons. Let $y_w$ and $y_l$ denote the chosen and rejected responses, respectively. These comparisons are modeled using the Bradley-Terry model:

\vspace{-10pt}
\begin{equation}
    p^*(y_w \succ y_l \mid x) = \sigma \big(r(x, y_w) - r(x, y_l)\big),
    \label{eq:bradley_terry}
\end{equation}
where $\sigma(t) = \frac{1}{1+e^{-t}}$ is the sigmoid function. By substituting \eqref{eq:reward_mapping}, the partition function $Z(x)$ cancels out, enabling DPO to directly compute the preference likelihoods from model probabilities without the need for explicit reward modeling.

\vspace{-10pt}
\begin{equation}
    r(x, y_w) - r(x, y_l) = \beta \log \frac{\pi^*(y_w|x)}{\pi_{\text{ref}}(y_w|x)} 
    - \beta \log \frac{\pi^*(y_l|x)}{\pi_{\text{ref}}(y_l|x)}.
    \label{eq:reward_difference}
\end{equation}

Finally, by maximizing the likelihood of observed preferences in the dataset $\mathcal{D}$, we derive the DPO loss:

\vspace{-10pt}
\begin{equation}
    \mathcal{L}_{\text{DPO}}(\pi; \pi_{\text{ref}}) = 
    -\mathbb{E}_{(x, y_w, y_l) \sim \mathcal{D}} \Biggl[
    \log \sigma \Big( 
    \beta \log \frac{\pi(y_w|x)}{\pi_{\text{ref}}(y_w|x)} 
    - \beta \log \frac{\pi(y_l|x)}{\pi_{\text{ref}}(y_l|x)}
    \Big) \Biggr].
    \label{eq:dpo_loss}
\end{equation}

This formulation directly aligns the policy $\pi$ with human preferences in a principled and efficient way, avoiding the instabilities and complexity associated with reinforcement learning.

\vspace{-3mm}
\section{METHODOLOGY}
\vspace{-3mm}

\begin{figure}[htp]
\centering
\includegraphics[width=\linewidth]{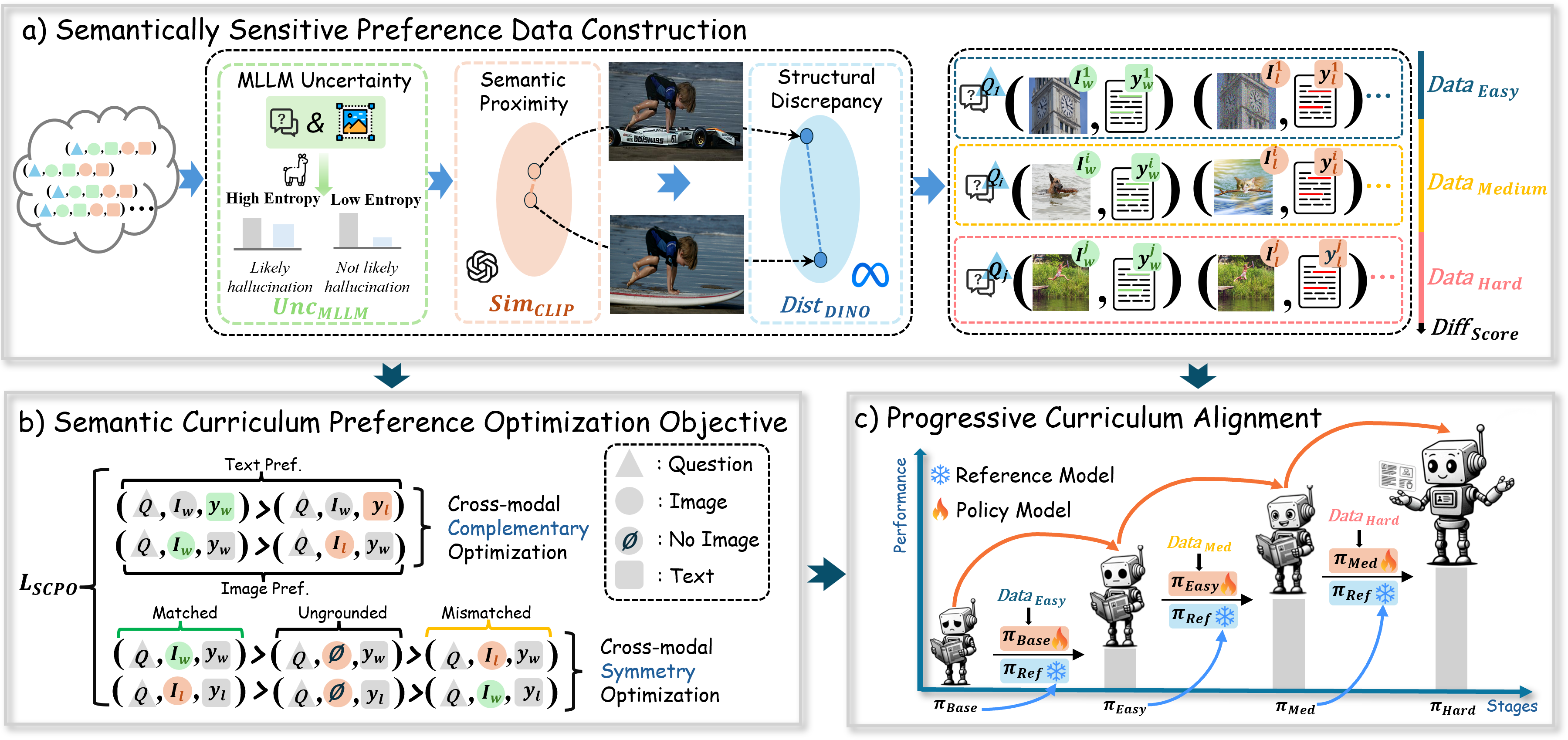}
\vspace{-5mm}
\caption{
    Overview of the SCPO framework. 
    \textbf{(a)} We first construct a preference dataset and partition it into an easy-to-hard curriculum based on semantic difficulty. 
    \textbf{(b)} The model is then trained using our symmetric, bidirectional SCPO objective. 
    \textbf{(c)} This training proceeds iteratively through the curriculum stages.
}
\label{fig:fig2}
\vspace{-5mm}
\end{figure}

Our proposed SCPO framework, illustrated in Figure~\ref{fig:fig2}, consists of three core components. We first detail the construction of our \textit{Semantic Curriculum Preference Pairs} (SCPP) dataset and its organization into a curriculum in Section~\ref{sec:scpp_construction}. We then introduce the symmetric SCPO objective in Section~\ref{sec:scpo_objective}. Finally, we describe the iterative alignment strategy with a dynamic reference model is described in Section~\ref{sec:iterative_learning}.

\vspace{-2mm}
\subsection{Semantically Sensitive Preference Data Construction}
\vspace{-2mm}

\label{sec:scpp_construction}
The construction of the SCPP dataset consists of two stages: (i) integration and refinement of public datasets, and (ii) curriculum-based partitioning into easy, medium, and hard sets.

\vspace{-2mm}
\subsubsection{Stage 1: Data Integration and Refinement}
\vspace{-2mm}

To enable fine-grained alignment, we construct the \textbf{SCPP} dataset, a large-scale collection where each sample consists of a prompt paired with two image-text pairs that differ by minimal but critical semantic attributes. The construction process integrates and refines several complementary public datasets. We leverage \textit{RLHF-V}~\citep{yu2024rlhf} for its fine-grained textual feedback and \textit{MVC}~\citep{wu2025symmetrical} for its minimally contrastive image pairs as the foundation of our dataset. To improve data quality and semantic diversity, we further incorporate \textit{HumanEdit}~\citep{bai2024humanedit} for its high-fidelity, human-guided visual edits, and \textit{Winoground}~\citep{thrush2022winoground} for its challenging compositional reasoning examples. To ensure that all sources adhere to the required format, we developed a processing pipeline to generate missing textual annotations where necessary. The quality of this automated process was further validated by manual spot-checks. A detailed description of each data source and our processing pipeline is provided in Appendix~\ref{app:data_details}.

\vspace{-2mm}
\subsubsection{Stage 2: Curriculum Partitioning}
\vspace{-2mm}

To enable a progressive learning trajectory, we partition the SCPP dataset into \textit{Easy}, \textit{Medium}, and \textit{Hard} subsets. This requires a principled method to quantify the difficulty of each preference pair. As illustrated in Figure~\ref{fig:fig2}(a), We propose a novel difficulty score taking into account three complementary factors: \textit{MLLM Uncertainty} ($\overline{\mathcal{H}}$), which captures the model's own confusion via semantic entropy; \textit{Semantic Proximity} ($s_{\mathrm{CLIP}}$), which measures high-level conceptual similarity using CLIP features~\citep{radford2021learning}; and \textit{Structural Discrepancy} ($d_{\mathrm{OT}}$), which quantifies fine-grained visual differences using DINOv2 features~\citep{oquab2024dinov2}. Detailed formulas for each metric are provided in Appendix~\ref{app:difficulty_metrics}.

We formulate the final difficulty score by aggregating the standardized values, or the $z$-scores, of these three metrics. The rationale is to identify pairs that are holistically challenging. A pair's difficulty increases with the model's own uncertainty, high $\overline{\mathcal{H}}$, since this indicates ambiguity. The difficulty is further amplified by high semantic proximity, high $s_{\mathrm{CLIP}}$, which creates strong conceptual distractors, and by high structural discrepancy, high $d_{\mathrm{OT}}$, which demands fine-grained visual perception. The combination of high semantic proximity and structural discrepancy is particularly challenging, as it forces the model to resolve conflicts between high-level semantics and low-level details~\citep{tong2024eyes}. By standardizing each metric to have zero mean and unit variance, we place them on a comparable scale. This allows us to combine them through a simple, unweighted sum, avoiding the need for additional hyperparameter tuning. Thus, this score prioritizes examples that are simultaneously ambiguous, semantically confusing, and structurally distinct. The final score is computed as:
\begin{equation}
\label{eq:difficulty_score}
\mathrm{Difficulty} = z_{\overline{\mathcal{H}}} + z_{s_{\mathrm{CLIP}}} + z_{d_{\mathrm{OT}}}.
\end{equation}
\paragraph{Dataset Partitioning.}
Based on the computed difficulty scores, we partition the SCPP dataset into three curriculum stages. The \textit{Easy} stage is composed of \textit{5k} pairs primarily derived from the RLHF-V dataset. These samples, whose negative examples often involve simple visual perturbations, consistently yield lower difficulty scores. We retain them as the foundational stage to build the model's basic factual grounding, leveraging their high-quality, fine-grained textual feedback. The \textit{Medium} stage is set to \textit{9k} pairs, a size determined empirically to yield the best balance of performance and data diversity, as detailed in our analysis in Appendix~\ref{app:data_ablation}. The remaining \textit{7k} pairs, which typically feature the most challenging semantic and structural conflicts, constitute the \textit{Hard} stage, designed to refine the model's capabilities on nuanced distinctions.

\vspace{-2mm}
\subsection{Semantic Curriculum Preference Optimization Objective}
\vspace{-2mm}

\label{sec:scpo_objective}
Our \textbf{SCPO Objective} is a novel loss function designed to enforce robust vision-language alignment. It is composed of two complementary components, which we detail below: \textit{Cross-modal Complementary Optimization (CCO)} and \textit{Cross-modal Symmetry Optimization (CSO)}.

\vspace{-2mm}
\paragraph{Cross-modal Complementary Optimization (CCO).}
\label{par:cco}
The CCO component learns preferences in both text and image modalities through two distinct objectives.

First, the \textbf{textual preference objective} trains the model to distinguish between correct and incorrect responses. Given a fixed image $I_w$, the model learns to prefer the chosen response $y_w$ over the rejected response $y_l$:
\begin{equation}
\mathcal{L}_{\text{text}} = 
-\log\sigma\!\left(
    \beta \left[
        \log\frac{\pi_\theta(y_w \mid I_w, Q)}{\pi_{\text{ref}}(y_w \mid I_w, Q)} -
        \log\frac{\pi_\theta(y_l \mid I_w, Q)}{\pi_{\text{ref}}(y_l \mid I_w, Q)}
    \right]
\right).
\label{eq:scpo_text_optimization}
\end{equation}

Second, the \textbf{visual preference objective} trains the model to identify the correct visual context. For a given correct response $y_w$, the model learns to prefer the matching image $I_w$ over the mismatched image $I_l$:
\begin{equation}
\mathcal{L}_{\text{image}} = 
-\log\sigma\!\left(
    \beta \left[
        \log\frac{\pi_\theta(y_w \mid I_w, Q)}{\pi_{\text{ref}}(y_w \mid I_w, Q)} -
        \log\frac{\pi_\theta(y_w \mid I_l, Q)}{\pi_{\text{ref}}(y_w \mid I_l, Q)}
    \right]
\right).
\label{eq:scpo_image_optimization}
\end{equation}

The total CCO loss is the sum of the objectives related to both textual and visual preferences:
\begin{equation}
\mathcal{L}_{\text{CCO}}\left(I_w, I_l, y_w, y_l, Q\right) = 
\mathcal{L}_{\text{text}}\left(y_w, y_l; I_w, Q\right) + 
\mathcal{L}_{\text{image}}\left(I_w, I_l; y_w, Q\right).
\label{eq:scpo_cco_optimization}
\end{equation}
This design embodies the complementary nature of the two objectives. The model learns a \textit{textual preference} for the correct response conditioned on the image, and simultaneously, a \textit{visual preference} for the correct image conditioned on the response. This complementary preference optimization fosters a more robust cross-modal alignment.

\vspace{-2mm}
\paragraph{Cross-modal Symmetry Optimization (CSO).}
\label{par:cso}
While CCO learns direct preferences, it does not explicitly prevent the model from ignoring the visual input. To enforce a stronger visual grounding and mitigate shortcut learning, we introduce the Cross-modal Symmetry Optimization (CSO) component, inspired by~\citep{wu2025symmetrical}. CSO operates on complete image-text pairs, rewarding consistent pairs while penalizing contradictory ones. This is achieved through two core objectives.

First, the \textbf{matching objective} encourages the model to prefer generating a response $y$ when conditioned on its matching image $I$, compared to generating it with no visual input (that is from language priors alone):
\begin{equation}
\mathcal{L}_{\text{match}}(I, y; Q) = 
-\log\sigma\left(
    \beta_1 \left[
        \log\frac{\pi_{\theta}(y \mid I, Q)}{\pi_{\text{ref}}(y \mid I, Q)} - 
        \log\frac{\pi_{\theta}(y \mid Q)}{\pi_{\text{ref}}(y \mid Q)}
    \right]
\right).
\label{eq:match_optimization}
\end{equation}

Second, the \textbf{contradiction objective} penalizes the model for associating a response $y$ with a mismatched image $I'$:
\begin{equation}
\mathcal{L}_{\text{contradict}}(I', y; Q) = 
-\log\sigma\left(
    \beta_2 \left[
        \log\frac{\pi_{\theta}(y \mid Q)}{\pi_{\text{ref}}(y \mid Q)} - 
        \log\frac{\pi_{\theta}(y \mid I', Q)}{\pi_{\text{ref}}(y \mid I', Q)}
    \right]
\right).
\label{eq:contradict_optimization}
\end{equation}
where $\beta_1$ and $\beta_2$ are scaling coefficients. The key insight of CSO is to apply these objectives symmetrically. The model is trained not only to match $(I_w, y_w)$ and reject the mismatched pair $(I_l, y_w)$, but also to match the pair $(I_l, y_l)$ and reject $(I_w, y_l)$. This symmetric application prevents the model from learning simple biases. The final CSO loss combines these four terms:
\begin{equation}
\begin{split}
\mathcal{L}_{\text{CSO}}\left(I_w, I_l, y_w, y_l, Q\right) = 
& \underbrace{\left( \mathcal{L}_{\text{match}}(I_w, y_w; Q) + \mathcal{L}_{\text{contradict}}(I_l, y_w; Q) \right)}_{\text{Alignment on } (I_w, y_w)} \\
& + \underbrace{\left( \mathcal{L}_{\text{match}}(I_l, y_l; Q) + \mathcal{L}_{\text{contradict}}(I_w, y_l; Q) \right)}_{\text{Alignment on }(I_l, y_l)}
\end{split}
\label{eq:scpo_cso_optimization}
\end{equation}

\paragraph{Total SCPO Objective.}
The final SCPO objective is a weighted sum of the CCO and CSO components, creating a unified loss function for robust vision-language alignment:
\begin{equation}
\mathcal{L}_{\mathrm{SCPO}} = 
\mathcal{L}_{\text{CCO}} + 
\lambda \, \mathcal{L}_{\text{CSO}}.
\label{eq:scpo_total_optimization}
\end{equation}
The hyperparameter $\lambda$ balances the influence of the CSO component.

\vspace{-2mm}
\subsection{Iterative Alignment with a Dynamic Reference Model}
\vspace{-2mm}
\label{sec:iterative_learning}

\begin{wrapfigure}{r}{0.55\textwidth}
    \vspace{-5pt} 
    \centering
    \includegraphics[width=\linewidth]{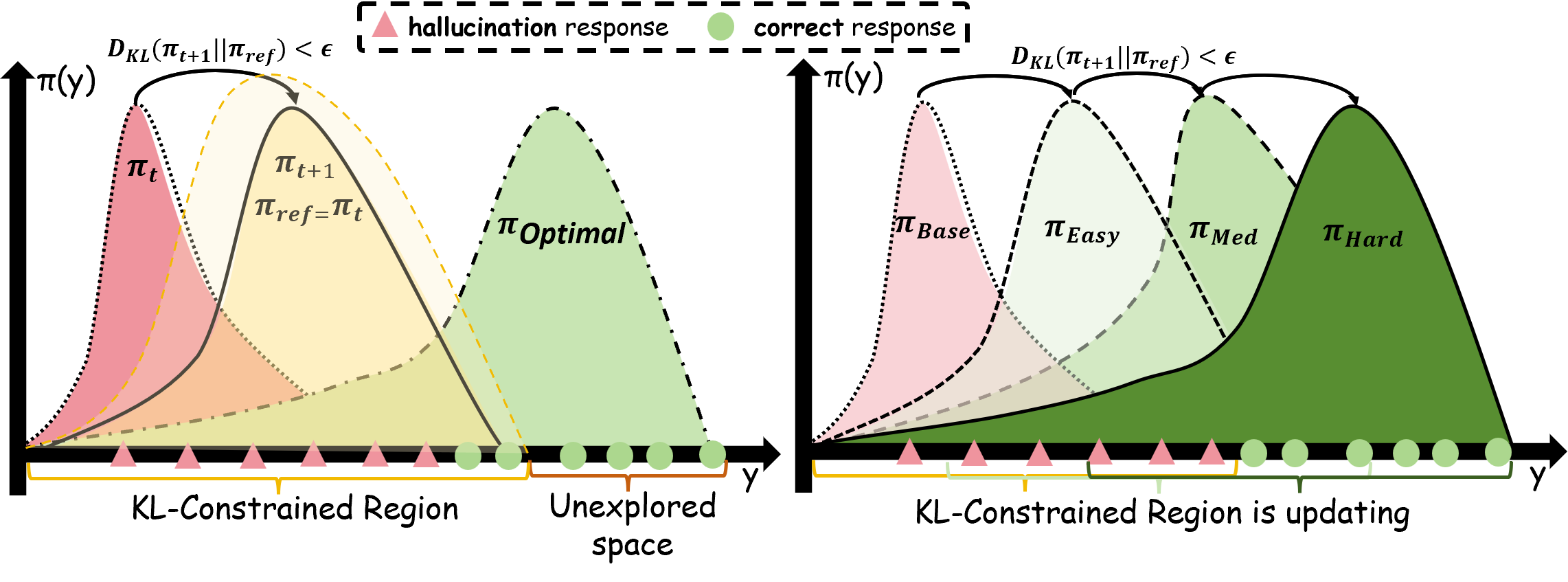} 
    \caption{
        Visualization of our iterative alignment. 
        \textit{(left)} In a single step, the policy update is constrained by the KL divergence from the reference $\pi_{\text{ref}}$. 
        \textit{(right)} Our dynamic approach updates this reference at each stage of the easy-to-hard curriculum, enabling a stable and cumulative learning trajectory.
    }
    \label{fig:iterative_alignment_viz}
    \vspace{-2mm}
\end{wrapfigure}

A static reference model in DPO can struggle with curriculum learning, as difficult, off-policy samples from later stages may cause optimization instability due to large KL penalties or vanishing gradients (Appendix~\ref{app:theoretical_analysis}). To address this, we introduce an iterative alignment strategy with a dynamic reference model, as visualized in Figure~\ref{fig:iterative_alignment_viz}. The training process consists of three sequential stages on the $\mathcal{D}_{\text{easy}}$, $\mathcal{D}_{\text{medium}}$, and $\mathcal{D}_{\text{hard}}$ subsets. After completing stage $t$ to produce an updated policy $\pi_t$, we dynamically reset the reference model for the subsequent stage $t+1$ by setting $\pi_{\text{ref}}^{(t+1)} \leftarrow \pi_t$.

This staged dynamic approach serves two critical functions. First, it mitigates the distribution shift problem common in DPO-based methods. By resetting the reference model, the KL divergence penalty is effectively reset to zero at the start of each new stage:
\begin{equation}
\label{eq:kl_reset}
\mathbb{E}_{x\sim\mathcal{D}_{t+1}} \left[ D_{\mathrm{KL}}\!\big(\pi_t(\cdot\mid x)\,\big\|\,\pi_{\text{ref}}^{(t+1)}(\cdot\mid x)\big) \right]
= \mathbb{E}_{x\sim\mathcal{D}_{t+1}} \left[ D_{\mathrm{KL}}(\pi_t\|\pi_t) \right] = 0.
\end{equation}
This prevents the policy from being overly constrained by a distant initial reference model, enabling more effective optimization. Second, this approach decomposes a single  complex optimization problem into a series of more manageable steps. Theoretically, the final policy $\pi_N$ after $N$ stages approximates the optimal policy for the cumulative reward: $\pi_N \propto \pi_0 \exp(\frac{1}{\beta}\sum_{t=1}^{N} r_t)$ (derivation in Appendix~\ref{sec:appendix_derivation}). This ensures a smooth and effective learning trajectory, as the policy is always regularized against a closely related and progressively improving reference model. The complete staged alignment process is summarized as follows:
\begin{equation}
\label{eq:iterative_flow}
\pi_{\text{Base}} 
\xrightarrow[\pi_{\text{ref}}=\pi_{\text{Base}}]{\mathcal{D}_{\text{easy}}, \mathcal{L}_{\text{SCPO}}} 
\pi_{\text{Easy}} 
\xrightarrow[\pi_{\text{ref}}=\pi_{\text{Easy}}]{\mathcal{D}_{\text{medium}}, \mathcal{L}_{\text{SCPO}}} 
\pi_{\text{Medium}} 
\xrightarrow[\pi_{\text{ref}}=\pi_{\text{Medium}}]{\mathcal{D}_{\text{hard}}, \mathcal{L}_{\text{SCPO}}} 
\pi_{\text{Hard}}
\end{equation}

\vspace{-3mm}
\section{EXPERIMENT AND RESULTS}
\vspace{-2mm}

\subsection{Experimental Setup}
\vspace{-2mm}

\paragraph{Implementation Details.}
We implement and evaluate our SCPO framework on three LLaVA base models: \textit{LLaVA-v1.5-7B}, \textit{LLaVA-v1.6-7B}, and \textit{LLaVA-v1.5-13B}~\citep{liu2023improved,liu2024llavanext}. All models utilize a CLIP ViT-L/14 vision encoder and a Vicuna LLM backbone. The alignment is performed on our \textit{SCPP} dataset, which is partitioned into three curriculum stages: \textit{Easy} (5k pairs), \textit{Medium} (9k pairs), and \textit{Hard} (7k pairs). We fine-tune all models for one epoch per stage with a batch size of \textit{32} and a learning rate of \textit{5e-7}. For our SCPO objective, we set the DPO coefficient $\beta = 0.1$ and the CSO coefficients $\beta_1 = \beta_2 = 0.1$. The weight of the CSO loss, $\lambda$, is set to \textit{0.2} to balance the magnitudes of the CCO and CSO loss terms, a value found to be robust in preliminary experiments (Appendix~\ref{app:lambda_sensitivity}). All experiments were conducted on 4 NVIDIA H20 GPUs.

\vspace{-2mm}
\paragraph{Evaluation Benchmarks.}
To provide a comprehensive evaluation, we assess the model performance on a suite of specialized hallucination benchmarks and general capability benchmarks..

\vspace{-2mm}
\subparagraph{Hallucination Benchmarks.}
We employ four benchmarks to measure hallucination mitigation. 
\textbf{1) Object HalBench}~\citep{rohrbach2018object}: A standard benchmark for assessing object hallucination. We evaluate across 300 instances, reporting hallucination rates at both the \textit{sentence} (\texttt{CHAIRs}↓) and \textit{object} levels (\texttt{CHAIRi}↓).
\textbf{2) AMBER-Generative}~\citep{wang2023llm}, a judgment-model-free benchmark for open-ended image description with \textit{1,004} samples, assesses factuality and coverage. We report five metrics: \textit{object hallucination} (\texttt{CHAIR}↓), \textit{object coverage} (\texttt{Cover}↑), \textit{response-level hallucination} (\texttt{HalRate}↓), \textit{cognitive hallucination} (\texttt{Cog}↓), and a composite \textit{F1 score} (\texttt{F1-Gen.}↑).\footnote{The \textit{F1-Gen.} score is not part of the original benchmark. We compute it as the harmonic mean of precision (1 - CHAIR) and recall (Cover) to provide a balanced metric for generative quality (Appendix~\ref{app:metric_details}).}
For \textbf{3) AMBER-Discriminative}~\citep{wang2023llm}, which evaluates binary verification queries across \textit{14,216} instances, we report overall \textit{Accuracy} (\texttt{Acc.}↑) and \textit{F1 score} (\texttt{F1-Dis.}↑).
\textbf{4) MMHal Bench}~\citep{sun2023aligning}: A question-answering benchmark with 96 images across 12 object categories. Following the official protocol, we use GPT-4 for evaluation and report the overall quality \textit{Score} (\texttt{Score}↑) and \textit{Hallucination Rate} (\texttt{HalRate}↓).

\vspace{-2mm}
\subparagraph{General Capability Benchmarks.}
To verify that our focus on factuality does not compromise general abilities, we also evaluate our method on two standard benchmarks. 
\textbf{1) LLaVA-in-the-Wild}~\citep{liu2024visual}: A benchmark with 60 open-ended questions to assess real-world conversational abilities. We report the average GPT-4-rated \textit{quality score} relative to the baseline (\texttt{Relative Score}↑).
\textbf{2) MMBench-CN}~\citep{liu2024mmbench}: A comprehensive multiple-choice benchmark. We report the overall \textit{accuracy} (\texttt{Acc.}↑).

\vspace{-3mm}
\paragraph{Baselines.}
We compare SCPO against a wide range of state-of-the-art models to demonstrate its effectiveness. Our comparison includes powerful general-purpose MLLMs such as \textit{GPT-4V}~\citep{gpt4V}, \textit{Qwen-VL-Chat}~\citep{bai2023qwenvl}, and \textit{MiniGemini}~\citep{li2024mini}. We also benchmark against a suite of leading alignment methods based on the DPO framework, all specifically designed to mitigate hallucinations. This includes \textit{Silkie}~\citep{li2024vlfeedback}, standard \textit{DPO}~\citep{rafailov2023direct} and its multimodal variants like \textit{mDPO}~\citep{wang2024mdpo}, as well as other advanced techniques such as \textit{EOS}~\citep{yue2024less}, \textit{HA-DPO}~\citep{zhao2023beyond}, \textit{HSA-DPO}~\citep{xiao2025detecting}, \textit{RLHF-V (HD)}~\citep{yu2024rlhf}, and \textit{HALVA}~\citep{sarkar2025mitigating}.

\vspace{-2mm}
\subsection{Main Results}
\vspace{-2mm}

\paragraph{Comparison with Leading MLLMs.}
We benchmark our SCPO-7B model against several powerful larger MLLMs in Table~\ref{table2_main_result_relative}. The results reveal a clear trade-off between generative breadth and factual accuracy. Although leading models like GPT-4V excel in metrics measuring descriptive coverage, our significantly smaller model demonstrates superior performance in hallucination mitigation. Notably, SCPO-7B reduces the \textit{CHAIRs} score by a relative \textit{48.5\%} compared to GPT-4V and achieves the highest accuracy on the \textit{AMBER-Discriminative} benchmark. This pattern suggests that our alignment method effectively enhances factual consistency, enabling a smaller open-source model to achieve a state-of-the-art level of reliability.

\vspace{-2mm}
\begin{table*}[h]
\caption{Performance comparison on hallucination benchmarks. Best results are in \textbf{bold}, second best are \underline{underlined}. Parenthesized numbers indicate relative change vs. GPT-4V. Results are cited from: $^\star$\citep{jiang2024modality}, $^\ddagger$\citet{xiao2025detecting}.}
\label{table2_main_result_relative}
\vspace{-2mm}
\resizebox{\textwidth}{!}{
\begin{tabular}{l|cc|ccccc|cc|cc}
\toprule
\multicolumn{1}{c}{\multirow{2}{*}{\textbf{Model}} } 
& \multicolumn{2}{c|}{\textbf{Object Hal}} 
& \multicolumn{5}{c|}{\textbf{AMBER-Generative}} 
& \multicolumn{2}{c|}{\textbf{AMBER-Discriminative}} 
& \multicolumn{2}{c}{\textbf{MMHal Bench}} \\ 
\cmidrule(lr){2-3}\cmidrule(lr){4-8}\cmidrule(lr){9-10}\cmidrule(lr){11-12}
~ & \textbf{CHAIRs↓} & \textbf{CHAIRi↓} 
& \textbf{CHAIR↓} & \textbf{Cover↑} & \textbf{HalRate↓} & \textbf{Cog↓} & \textbf{F1-Gen.↑} 
& \textbf{Acc.↑} & \textbf{F1-Dis.↑} 
& \textbf{Score↑} & \textbf{HalRate↓} \\ 
\midrule

\multicolumn{1}{l|}{GPT-4V $^\star$}        
& 13.6 & \multicolumn{1}{c|}{7.3} 
& \underline{4.6} & \textbf{67.1} & 30.7 & 2.6 & \multicolumn{1}{c|}{\textbf{78.8}}
& \underline{83.4} & \multicolumn{1}{c|}{87.4} 
& \textbf{3.49} & \textbf{0.28} \\ 
\hdashline

\multicolumn{1}{l|}{Qwen-VL-Chat-10B $^\ddagger$} 
& 36.0\textbf{\color[HTML]{E41A1C}\scriptsize{(↑22.4)}} 
& 21.3\textbf{\color[HTML]{E41A1C}\scriptsize{(↑14.0)}} 
& 6.6\textbf{\color[HTML]{E41A1C}\scriptsize{(↑2.0)}} 
& 53.2\textbf{\color[HTML]{E41A1C}\scriptsize{(↓13.9)}} 
& 31.0\textbf{\color[HTML]{E41A1C}\scriptsize{(↑0.3)}} 
& 2.9\textbf{\color[HTML]{E41A1C}\scriptsize{(↑0.3)}} 
& - 
& 81.9\textbf{\color[HTML]{E41A1C}\scriptsize{(↓1.5)}} 
& 86.4\textbf{\color[HTML]{E41A1C}\scriptsize{(↓1.0)}} 
& 2.89\textbf{\color[HTML]{E41A1C}\scriptsize{(↓0.60)}} 
& 0.43\textbf{\color[HTML]{E41A1C}\scriptsize{(↑0.15)}} \\ 

\multicolumn{1}{l|}{Silkie-10B $^\ddagger$}       
& 25.3\textbf{\color[HTML]{E41A1C}\scriptsize{(↑11.7)}} 
& 13.9\textbf{\color[HTML]{E41A1C}\scriptsize{(↑6.6)}} 
& 5.4\textbf{\color[HTML]{E41A1C}\scriptsize{(↑0.8)}} 
& 55.8\textbf{\color[HTML]{E41A1C}\scriptsize{(↓11.3)}} 
& \underline{29.0}\textbf{\color[HTML]{4DAF4A}\scriptsize{(↓1.7)}} 
& \underline{2.0}\textbf{\color[HTML]{4DAF4A}\scriptsize{(↓0.6)}} 
& - 
& 82.2\textbf{\color[HTML]{E41A1C}\scriptsize{(↓1.2)}} 
& \underline{87.6}\textbf{\color[HTML]{4DAF4A}\scriptsize{(↑0.2)}} 
& 3.01\textbf{\color[HTML]{E41A1C}\scriptsize{(↓0.48)}} 
& 0.41\textbf{\color[HTML]{E41A1C}\scriptsize{(↑0.13)}} \\ 

\multicolumn{1}{l|}{InstructBLIP-13B $^\ddagger$} 
& 25.9\textbf{\color[HTML]{E41A1C}\scriptsize{(↑12.3)}} 
& 14.3\textbf{\color[HTML]{E41A1C}\scriptsize{(↑7.0)}} 
& 8.8\textbf{\color[HTML]{E41A1C}\scriptsize{(↑4.2)}} 
& 52.2\textbf{\color[HTML]{E41A1C}\scriptsize{(↓14.9)}} 
& 38.2\textbf{\color[HTML]{E41A1C}\scriptsize{(↑7.5)}} 
& 4.4\textbf{\color[HTML]{E41A1C}\scriptsize{(↑1.8)}} 
& - & - & - 
& 2.14\textbf{\color[HTML]{E41A1C}\scriptsize{(↓1.35)}} 
& 0.58\textbf{\color[HTML]{E41A1C}\scriptsize{(↑0.30)}} \\ 

\multicolumn{1}{l|}{MiniGemini-34B $^\star$}    
& 14.5\textbf{\color[HTML]{E41A1C}\scriptsize{(↑0.9)}} 
& 8.0\textbf{\color[HTML]{E41A1C}\scriptsize{(↑0.7)}} 
& - & - & - & - & - 
& 82.6\textbf{\color[HTML]{E41A1C}\scriptsize{(↓0.8)}} 
& \underline{87.6}\textbf{\color[HTML]{4DAF4A}\scriptsize{(↑0.2)}} 
& 3.08\textbf{\color[HTML]{E41A1C}\scriptsize{(↓0.41)}} 
& 0.38\textbf{\color[HTML]{E41A1C}\scriptsize{(↑0.10)}} \\ 

\hdashline

\multicolumn{1}{l|}{\textbf{LLaVA-v1.6-7B $^\S$}} 
& \underline{12.0}\textbf{\color[HTML]{4DAF4A}\scriptsize{(↓1.6)}} 
& \underline{6.8}\textbf{\color[HTML]{4DAF4A}\scriptsize{(↓0.5)}} 
& 8.7\textbf{\color[HTML]{E41A1C}\scriptsize{(↑4.1)}} 
& \underline{61.1}\textbf{\color[HTML]{E41A1C}\scriptsize{(↓6.0)}} 
& 49.7\textbf{\color[HTML]{E41A1C}\scriptsize{(↑19.0)}} 
& 4.2\textbf{\color[HTML]{E41A1C}\scriptsize{(↑1.6)}} 
& 73.2\textbf{\color[HTML]{E41A1C}\scriptsize{(↓5.6)}} 
& 82.8\textbf{\color[HTML]{E41A1C}\scriptsize{(↓0.6)}} 
& 87.0\textbf{\color[HTML]{E41A1C}\scriptsize{(↓0.4)}} 
& 2.80\textbf{\color[HTML]{E41A1C}\scriptsize{(↓0.69)}} 
& 0.43\textbf{\color[HTML]{E41A1C}\scriptsize{(↑0.15)}} \\ 

\rowcolor[HTML]{ECF4FF} 
\multicolumn{1}{l|}{\textbf{   +SCPO-7B (ours)}}           
& \textbf{7.0}\textbf{\color[HTML]{4DAF4A}\scriptsize{(↓6.6)}} 
& \multicolumn{1}{c|}{\textbf{4.4}\textbf{\color[HTML]{4DAF4A}\scriptsize{(↓2.9)}}} 
& \textbf{4.5}\textbf{\color[HTML]{4DAF4A}\scriptsize{(↓0.1)}} 
& 60.2\textbf{\color[HTML]{E41A1C}\scriptsize{(↓6.9)}} 
& \textbf{27.1}\textbf{\color[HTML]{4DAF4A}\scriptsize{(↓3.6)}} 
& \textbf{1.9}\textbf{\color[HTML]{4DAF4A}\scriptsize{(↓0.7)}} 
& \multicolumn{1}{c|}{\underline{73.8}\textbf{\color[HTML]{E41A1C}\scriptsize{(↓5.0)}}}
& \textbf{85.4}\textbf{\color[HTML]{4DAF4A}\scriptsize{(↑2.0)}} 
& \multicolumn{1}{c|}{\textbf{89.2}\textbf{\color[HTML]{4DAF4A}\scriptsize{(↑1.8)}}} 
& \underline{3.16}\textbf{\color[HTML]{E41A1C}\scriptsize{(↓0.33)}} 
& \underline{0.32}\textbf{\color[HTML]{E41A1C}\scriptsize{(↑0.04)}} \\ 
\bottomrule
\end{tabular}
}
\vspace{-4mm}
\end{table*}

\paragraph{Comparison with Alignment Baselines.}

We compare the SCPO framework with several alignment baselines across various LLaVA architectures, as shown in Table~\ref{table2_main_result3}. The results highlight that SCPO consistently outperforms prior methods of all model sizes and nearly all hallucination-related metrics. To ensure a fair comparison, we include \textit{DPO (SCPP)}, which applies the standard DPO objective to our data and training methods. While \textit{DPO (SCPP)} yields competitive results, SCPO provides a substantial further improvement, reducing the \textit{CHAIRs score} by 40.3\% and the \textit{AMBER HalRate} by 36.3\% compared to \textit{DPO (SCPP)}. This confirms that the combination of our data, progressive curriculum, and objectives is essential for achieving state-of-the-art hallucination mitigation.

\vspace{-2mm}
\begin{table*}[h]
    \caption{Performance of SCPO against alignment baselines on various LLaVA models. Our method's performance is reported after each progressive curriculum stage. DPO (SCPP) denotes the standard DPO trained with our data and method. Baselines marked with $\S$ are our re-evaluations; others are cited from their respective papers: $\star$~\citep{wang2024mdpo}, $\ddagger$~\citep{sarkarmitigating}, $\dagger$~\citep{fu2025chip}, $\sharp$~\citep{yang2025mitigating}. Best and second-best results are in \textbf{bold} and \underline{underlined}. Parentheses show absolute change from the corresponding baseline.}
    \label{table2_main_result3}
    \vspace{-1mm}
    \resizebox{\textwidth}{!}{
    \begin{tabular}{l|cc|ccccc|cc|cc}
    \toprule
    \multicolumn{1}{c}{\multirow{2}{*}{\textbf{Algorithm}}} & \multicolumn{2}{c}{\textbf{Object Hal}} & \multicolumn{5}{c}{ \textbf{AMBER-Generative}} & \multicolumn{2}{c}{\textbf{AMBER-Discriminative}} & \multicolumn{2}{c}{\textbf{MMHal Bench}} \\ \cmidrule(lr){2-3}\cmidrule(lr){4-8}\cmidrule(lr){9-10}\cmidrule(lr){11-12}
    
    ~ & {\textbf{CHAIRs↓}} & {\textbf{CHAIRi↓}} & {\textbf{CHAIR↓}} & { \textbf{Cover↑}} & {\textbf{HalRate↓}} & {{ \textbf{Cog↓}}} & { \textbf{F1-Gen.↑}} & {\textbf{Acc.↑}} &{{\textbf{F1-Dis.↑}}} & { \textbf{Score↑}} & { \textbf{HalRate↓}} \\ \midrule
    
    
    \multicolumn{1}{l|}{\textbf{LLaVA-v1.5-7B $^\S$}} & 54.7\textbf{\color[HTML]{FFFFFF}} & 26.5\textbf{\color[HTML]{FFFFFF}} & 8.8\textbf{\color[HTML]{FFFFFF}} & 50.1\textbf{\color[HTML]{FFFFFF}} & 40.4\textbf{\color[HTML]{FFFFFF}} & 4.7\textbf{\color[HTML]{FFFFFF}} & 64.7\textbf{\color[HTML]{FFFFFF}} & 80.8\textbf{\color[HTML]{FFFFFF}}  & 85.2\textbf{\color[HTML]{FFFFFF}} & 2.18\textbf{\color[HTML]{FFFFFF}} & 0.59\textbf{\color[HTML]{FFFFFF}} \\
    +DPO $^\star$ & 49.0\textbf{\color[HTML]{FFFFFF}} & 13.0\textbf{\color[HTML]{FFFFFF}} & 6.5\textbf{\color[HTML]{FFFFFF}} & \underline{55.5}\textbf{\color[HTML]{FFFFFF}} & 34.5\textbf{\color[HTML]{FFFFFF}} & 2.3\textbf{\color[HTML]{FFFFFF}} & \underline{69.7}\textbf{\color[HTML]{FFFFFF}} & -\textbf{\color[HTML]{FFFFFF}} & -\textbf{\color[HTML]{FFFFFF}} & 2.14\textbf{\color[HTML]{FFFFFF}} & 0.65\textbf{\color[HTML]{FFFFFF}} \\ 
    +mDPO $^\star$ & 35.7\textbf{\color[HTML]{FFFFFF}} & \textbf{9.8}\textbf{\color[HTML]{FFFFFF}} & 4.4\textbf{\color[HTML]{FFFFFF}} & 52.4\textbf{\color[HTML]{FFFFFF}} & 24.5\textbf{\color[HTML]{FFFFFF}} & 2.4\textbf{\color[HTML]{FFFFFF}} & 67.7\textbf{\color[HTML]{FFFFFF}} & -\textbf{\color[HTML]{FFFFFF}} & -\textbf{\color[HTML]{FFFFFF}} & 2.39\textbf{\color[HTML]{FFFFFF}} & 0.54 \textbf{\color[HTML]{FFFFFF}}\\ 
    +EOS $^\ddagger$ & 40.2\textbf{\color[HTML]{FFFFFF}} & 12.3\textbf{\color[HTML]{FFFFFF}} & 5.1\textbf{\color[HTML]{FFFFFF}} & 49.1\textbf{\color[HTML]{FFFFFF}} & 22.7\textbf{\color[HTML]{FFFFFF}} & 2.0\textbf{\color[HTML]{FFFFFF}} & 64.7\textbf{\color[HTML]{FFFFFF}} & -\textbf{\color[HTML]{FFFFFF}} & 75.6\textbf{\color[HTML]{FFFFFF}} & 2.03\textbf{\color[HTML]{FFFFFF}} & 0.59 \textbf{\color[HTML]{FFFFFF}}\\
    +HA-DPO $^\ddagger$ & 38.2\textbf{\color[HTML]{FFFFFF}} & \underline{11.0}\textbf{\color[HTML]{FFFFFF}} & 6.7\textbf{\color[HTML]{FFFFFF}} & 49.8\textbf{\color[HTML]{FFFFFF}} & 30.9\textbf{\color[HTML]{FFFFFF}} & 3.3\textbf{\color[HTML]{FFFFFF}} & 64.9\textbf{\color[HTML]{FFFFFF}} & -\textbf{\color[HTML]{FFFFFF}} & 78.1\textbf{\color[HTML]{FFFFFF}} & 1.97\textbf{\color[HTML]{FFFFFF}} & 0.59 \textbf{\color[HTML]{FFFFFF}}\\
    +HALVA $^\S$ & 52.7\textbf{\color[HTML]{FFFFFF}} & 26.1\textbf{\color[HTML]{FFFFFF}} & 8.3\textbf{\color[HTML]{FFFFFF}} & 51.0\textbf{\color[HTML]{FFFFFF}} & 37.1\textbf{\color[HTML]{FFFFFF}} & 4.2\textbf{\color[HTML]{FFFFFF}} & 65.5\textbf{\color[HTML]{FFFFFF}} & 81.1 \textbf{\color[HTML]{FFFFFF}} & 85.5\textbf{\color[HTML]{FFFFFF}} & 2.19\textbf{\color[HTML]{FFFFFF}} & 0.58 \textbf{\color[HTML]{FFFFFF}}\\
    +DPO (SCPP) & 34.0\textbf{\color[HTML]{FFFFFF}} & 18.7\textbf{\color[HTML]{FFFFFF}} & 4.9\textbf{\color[HTML]{FFFFFF}} & \textbf{58.2}\textbf{\color[HTML]{FFFFFF}} & 30.6\textbf{\color[HTML]{FFFFFF}} & 1.8\textbf{\color[HTML]{FFFFFF}} & \textbf{72.2}\textbf{\color[HTML]{FFFFFF}} & 81.5\textbf{\color[HTML]{FFFFFF}} & \underline{86.8}\textbf{\color[HTML]{FFFFFF}} & 2.37\textbf{\color[HTML]{FFFFFF}} & 0.52\textbf{\color[HTML]{FFFFFF}} \\
    \cellcolor[HTML]{ECF4FF}+\textbf{Easy(ours)} & \cellcolor[HTML]{ECF4FF}27.7\textbf{\color[HTML]{4DAF4A}\scriptsize{(↓27.0)}} & \cellcolor[HTML]{ECF4FF}15.0\textbf{\color[HTML]{4DAF4A}\scriptsize{(↓11.5)}} & \cellcolor[HTML]{ECF4FF}4.0\textbf{\color[HTML]{4DAF4A}\scriptsize{(↓4.8)}} & \cellcolor[HTML]{ECF4FF}50.7\textbf{\color[HTML]{4DAF4A}\scriptsize{(↑0.6)}} & \cellcolor[HTML]{ECF4FF}21.7\textbf{\color[HTML]{4DAF4A}\scriptsize{(↓18.7)}} & \cellcolor[HTML]{ECF4FF}2.0\textbf{\color[HTML]{4DAF4A}\scriptsize{(↓2.7)}} & \cellcolor[HTML]{ECF4FF}66.4\textbf{\color[HTML]{4DAF4A}\scriptsize{(↑1.7)}} & \cellcolor[HTML]{ECF4FF}79.4\textbf{\color[HTML]{FB8072}\scriptsize{(↓1.4)}}  & \cellcolor[HTML]{ECF4FF}85.6\textbf{\color[HTML]{4DAF4A}\scriptsize{(↑0.4)}} & \cellcolor[HTML]{ECF4FF}2.37\textbf{\color[HTML]{4DAF4A}\scriptsize{(↑0.19)}} & \cellcolor[HTML]{ECF4FF}0.54\textbf{\color[HTML]{4DAF4A}\scriptsize{(↓0.05)}}\\
    \cellcolor[HTML]{ECF4FF}+\textbf{Medium(ours)} & \cellcolor[HTML]{ECF4FF}\underline{25.3}\textbf{\color[HTML]{4DAF4A}\scriptsize{(↓29.4)}} & \cellcolor[HTML]{ECF4FF}13.3\textbf{\color[HTML]{4DAF4A}\scriptsize{(↓13.2)}} & \cellcolor[HTML]{ECF4FF}\underline{3.5}\textbf{\color[HTML]{4DAF4A}\scriptsize{(↓5.3)}} & \cellcolor[HTML]{ECF4FF}52.8\textbf{\color[HTML]{4DAF4A}\scriptsize{(↑2.7)}} & \cellcolor[HTML]{ECF4FF}\underline{21.3}\textbf{\color[HTML]{4DAF4A}\scriptsize{(↓19.1)}} & \cellcolor[HTML]{ECF4FF}\underline{1.7}\textbf{\color[HTML]{4DAF4A}\scriptsize{(↓3.0)}} & 
    \cellcolor[HTML]{ECF4FF}68.3\textbf{\color[HTML]{4DAF4A}\scriptsize{(↑3.6)}} & \cellcolor[HTML]{ECF4FF}\underline{82.9}\textbf{\color[HTML]{4DAF4A}\scriptsize{(↑2.1)}}  & \cellcolor[HTML]{ECF4FF}\textbf{87.3}\textbf{\color[HTML]{4DAF4A}\scriptsize{(↑2.1)}} & \cellcolor[HTML]{ECF4FF}\textbf{2.73}\textbf{\color[HTML]{4DAF4A}\scriptsize{(↑0.55)}} & \cellcolor[HTML]{ECF4FF}\underline{0.48}\textbf{\color[HTML]{4DAF4A}\scriptsize{(↓0.11)}}\\
    \cellcolor[HTML]{ECF4FF}+\textbf{Hard(ours)} & \cellcolor[HTML]{ECF4FF}\textbf{20.3}\textbf{\color[HTML]{4DAF4A}\scriptsize{(↓34.4)}} & \cellcolor[HTML]{ECF4FF}11.6\textbf{\color[HTML]{4DAF4A}\scriptsize{(↓14.9)}} & \cellcolor[HTML]{ECF4FF}\textbf{3.3\scriptsize{(↓5.5)}} & \cellcolor[HTML]{ECF4FF}53.6\textbf{\color[HTML]{4DAF4A}\scriptsize{(↑3.5)}} & \cellcolor[HTML]{ECF4FF}\textbf{19.5}\textbf{\color[HTML]{4DAF4A}\scriptsize{(↓20.9)}} & \cellcolor[HTML]{ECF4FF}\textbf{1.4}\textbf{\color[HTML]{4DAF4A}\scriptsize{(↓3.3)}} & \cellcolor[HTML]{ECF4FF}69.0\textbf{\color[HTML]{4DAF4A}\scriptsize{(↑4.3)}} & \cellcolor[HTML]{ECF4FF}\textbf{83.2}\textbf{\color[HTML]{4DAF4A}\scriptsize{(↑2.4)}} & \cellcolor[HTML]{ECF4FF}\textbf{87.3}\textbf{\color[HTML]{4DAF4A}\scriptsize{(↑2.1)}} & \cellcolor[HTML]{ECF4FF}\underline{2.47}\textbf{\color[HTML]{4DAF4A}\scriptsize{(↑0.29)}} & \cellcolor[HTML]{ECF4FF}\textbf{0.44}\textbf{\color[HTML]{4DAF4A}\scriptsize{(↓0.15)}}\\
    
    \midrule
    
    \multicolumn{1}{l|}{\textbf{LLaVA-v1.6-7B $^\S$}} & 12.0\textbf{\color[HTML]{FFFFFF}} & 6.8\textbf{\color[HTML]{FFFFFF}} & 8.7\textbf{\color[HTML]{FFFFFF}} & \underline{61.1}\textbf{\color[HTML]{FFFFFF}} & 49.7\textbf{\color[HTML]{FFFFFF}} & 4.2\textbf{\color[HTML]{FFFFFF}} & 73.2\textbf{\color[HTML]{FFFFFF}} & 82.8\textbf{\color[HTML]{FFFFFF}} & 87.0\textbf{\color[HTML]{FFFFFF}} & 2.80\textbf{\color[HTML]{FFFFFF}} & 0.43 \textbf{\color[HTML]{FFFFFF}}\\
    +DPO $^\dagger$ & 11.0\textbf{\color[HTML]{FFFFFF}} & 6.6\textbf{\color[HTML]{FFFFFF}} & 5.9\textbf{\color[HTML]{FFFFFF}} & 61.0\textbf{\color[HTML]{FFFFFF}} & 38.9\textbf{\color[HTML]{FFFFFF}} & 3.0\textbf{\color[HTML]{FFFFFF}} & \underline{74.0}\textbf{\color[HTML]{FFFFFF}} & -\textbf{\color[HTML]{FFFFFF}} & 87.4\textbf{\color[HTML]{FFFFFF}} & 2.70\textbf{\color[HTML]{FFFFFF}} & 0.44 \textbf{\color[HTML]{FFFFFF}}\\
    +DPO (SCPP) & 10.7\textbf{\color[HTML]{FFFFFF}} & 5.9\textbf{\color[HTML]{FFFFFF}} & 6.7\textbf{\color[HTML]{FFFFFF}} & \textbf{62.5}\textbf{\color[HTML]{FFFFFF}} & 37.4\textbf{\color[HTML]{FFFFFF}} & 2.9\textbf{\color[HTML]{FFFFFF}} & \textbf{74.9}\textbf{\color[HTML]{FFFFFF}} & 85.0\textbf{\color[HTML]{FFFFFF}} & 88.3\textbf{\color[HTML]{FFFFFF}} & 2.55\textbf{\color[HTML]{FFFFFF}} & 0.53\textbf{\color[HTML]{FFFFFF}} \\
    \cellcolor[HTML]{ECF4FF}+\textbf{Easy(ours)} & \cellcolor[HTML]{ECF4FF}\underline{8.3}\textbf{\color[HTML]{4DAF4A}\scriptsize{(↓3.7)}} & \cellcolor[HTML]{ECF4FF}\underline{5.3}\textbf{\color[HTML]{4DAF4A}\scriptsize{(↓1.48)}} & \cellcolor[HTML]{ECF4FF}\textbf{4.4}\textbf{\color[HTML]{4DAF4A}\scriptsize{(↓4.3)}} & \cellcolor[HTML]{ECF4FF}59.3\textbf{\color[HTML]{FB8072}\scriptsize{(↓1.8)}} & \cellcolor[HTML]{ECF4FF}\underline{28.8}\textbf{\color[HTML]{4DAF4A}\scriptsize{(↓20.9)}} & \cellcolor[HTML]{ECF4FF}\underline{2.1}\textbf{\color[HTML]{4DAF4A}\scriptsize{(↓2.1)}} & \cellcolor[HTML]{ECF4FF}73.2\textbf{\color[HTML]{ECF4FF}} & \cellcolor[HTML]{ECF4FF}82.3\textbf{\color[HTML]{FB8072}\scriptsize{(↓0.5)}} & \cellcolor[HTML]{ECF4FF}87.3\textbf{\color[HTML]{4DAF4A}\scriptsize{(↑0.3)}} & \cellcolor[HTML]{ECF4FF}2.64\textbf{\color[HTML]{FB8072}\scriptsize{(↓0.16)}} & \cellcolor[HTML]{ECF4FF}0.46\textbf{\color[HTML]{FB8072}\scriptsize{(↑0.03)}}\\
    \cellcolor[HTML]{ECF4FF}+\textbf{Medium(ours)} & \cellcolor[HTML]{ECF4FF}{8.6}\textbf{\color[HTML]{4DAF4A}\scriptsize{(↓3.4)}} & \cellcolor[HTML]{ECF4FF}\underline{5.3}\textbf{\color[HTML]{4DAF4A}\scriptsize{(↓1.51)}} & \cellcolor[HTML]{ECF4FF}{4.8}\textbf{\color[HTML]{4DAF4A}\scriptsize{(↓4.2)}} & \cellcolor[HTML]{ECF4FF}60.3\textbf{\color[HTML]{FB8072}\scriptsize{(↓0.8)}} & \cellcolor[HTML]{ECF4FF}31.2\textbf{\color[HTML]{4DAF4A}\scriptsize{(↓18.5)}} & \cellcolor[HTML]{ECF4FF}2.2\textbf{\color[HTML]{4DAF4A}\scriptsize{(↓2.0)}} & 
    \cellcolor[HTML]{ECF4FF}73.8\textbf{\color[HTML]{4DAF4A}\scriptsize{(↑0.62)}} & \cellcolor[HTML]{ECF4FF}\underline{85.1}\textbf{\color[HTML]{4DAF4A}\scriptsize{(↑2.3)}} & \cellcolor[HTML]{ECF4FF}\underline{88.9}\textbf{\color[HTML]{4DAF4A}\scriptsize{(↑1.9)}} & \cellcolor[HTML]{ECF4FF}\underline{2.94}\textbf{\color[HTML]{4DAF4A}\scriptsize{(↑0.14)}} & \cellcolor[HTML]{ECF4FF}\underline{0.38}\textbf{\color[HTML]{4DAF4A}\scriptsize{(↓0.05)}}\\
    \cellcolor[HTML]{ECF4FF}+\textbf{Hard(ours)} & \cellcolor[HTML]{ECF4FF}\textbf{7.0}\textbf{\color[HTML]{4DAF4A}\scriptsize{(↓5.0)}} & \cellcolor[HTML]{ECF4FF}\textbf{4.4}\textbf{\color[HTML]{4DAF4A}\scriptsize{(↓2.37)}} & \cellcolor[HTML]{ECF4FF}\underline{4.5}\textbf{\color[HTML]{4DAF4A}\scriptsize{(↓4.2)}} & \cellcolor[HTML]{ECF4FF}60.2\textbf{\color[HTML]{FB8072}\scriptsize{(↓0.9)}} & \cellcolor[HTML]{ECF4FF}\textbf{27.1}\textbf{\color[HTML]{4DAF4A}\scriptsize{(↓22.6)}} & \cellcolor[HTML]{ECF4FF}\textbf{1.9}\textbf{\color[HTML]{4DAF4A}\scriptsize{(↓2.3)}} & \cellcolor[HTML]{ECF4FF}73.8\textbf{\color[HTML]{4DAF4A}\scriptsize{(↑0.64)}} & \cellcolor[HTML]{ECF4FF}\textbf{85.4}\textbf{\color[HTML]{4DAF4A}\scriptsize{(↑2.6)}} & \cellcolor[HTML]{ECF4FF}\textbf{89.2}\textbf{\color[HTML]{4DAF4A}\scriptsize{(↑2.2)}} & \cellcolor[HTML]{ECF4FF}\textbf{3.16}\textbf{\color[HTML]{4DAF4A}\scriptsize{(↑0.36)}} & \cellcolor[HTML]{ECF4FF}\textbf{0.32}\textbf{\color[HTML]{4DAF4A}\scriptsize{(↓0.11)}}\\
    
    \midrule
    
    \multicolumn{1}{l|}{\textbf{LLaVA-v1.5-13B $^\S$}} & 49.3\textbf{\color[HTML]{FFFFFF}} & 23.9\textbf{\color[HTML]{FFFFFF}} & 8.8\textbf{\color[HTML]{FFFFFF}} & 50.3\textbf{\color[HTML]{FFFFFF}} & 37.2\textbf{\color[HTML]{FFFFFF}} & 4.3\textbf{\color[HTML]{FFFFFF}} & 64.8\textbf{\color[HTML]{FFFFFF}} & 83.5\textbf{\color[HTML]{FFFFFF}} & 86.6\textbf{\color[HTML]{FFFFFF}} & 2.31\textbf{\color[HTML]{FFFFFF}} & 0.55 \textbf{\color[HTML]{FFFFFF}}\\
    +HSA-DPO $^\sharp$ & -\textbf{\color[HTML]{FFFFFF}} & -\textbf{\color[HTML]{FFFFFF}} & \textbf{2.1}\textbf{\color[HTML]{FFFFFF}} & 47.3\textbf{\color[HTML]{FFFFFF}} & \textbf{13.4}\textbf{\color[HTML]{FFFFFF}} & \textbf{1.2}\textbf{\color[HTML]{FFFFFF}} & 63.8\textbf{\color[HTML]{FFFFFF}} & -\textbf{\color[HTML]{FFFFFF}} & -\textbf{\color[HTML]{FFFFFF}} & 2.61\textbf{\color[HTML]{FFFFFF}} & 0.48\textbf{\color[HTML]{FFFFFF}}\\
    +RLHF-V (HD) $^\sharp$ & -\textbf{\color[HTML]{FFFFFF}} & -\textbf{\color[HTML]{FFFFFF}} & 6.3\textbf{\color[HTML]{FFFFFF}} & 46.1\textbf{\color[HTML]{FFFFFF}} & 25.1\textbf{\color[HTML]{FFFFFF}} & 2.1\textbf{\color[HTML]{FFFFFF}} & 61.8\textbf{\color[HTML]{FFFFFF}} & -\textbf{\color[HTML]{FFFFFF}} & -\textbf{\color[HTML]{FFFFFF}} & \underline{2.81}\textbf{\color[HTML]{FFFFFF}} & 0.49 \textbf{\color[HTML]{FFFFFF}}\\
    +HALVA $^\S$ & 47.0\textbf{\color[HTML]{FFFFFF}} & 22.9\textbf{\color[HTML]{FFFFFF}} & 8.4\textbf{\color[HTML]{FFFFFF}} & 50.5\textbf{\color[HTML]{FFFFFF}} & 35.4\textbf{\color[HTML]{FFFFFF}} & 4.0\textbf{\color[HTML]{FFFFFF}} & 65.1\textbf{\color[HTML]{FFFFFF}} & 85.2\textbf{\color[HTML]{FFFFFF}} & 88.3\textbf{\color[HTML]{FFFFFF}} & {2.44}\textbf{\color[HTML]{FFFFFF}} & 0.52 \textbf{\color[HTML]{FFFFFF}}\\
    +DPO (SCPP) & 34.0\textbf{\color[HTML]{FFFFFF}} & 17.5\textbf{\color[HTML]{FFFFFF}} & 4.9\textbf{\color[HTML]{FFFFFF}} & \textbf{55.7}\textbf{\color[HTML]{FFFFFF}} & 27.9\textbf{\color[HTML]{FFFFFF}} & 1.8\textbf{\color[HTML]{FFFFFF}} & \textbf{70.3}\textbf{\color[HTML]{FFFFFF}} & 86.2\textbf{\color[HTML]{FFFFFF}} & 82.8\textbf{\color[HTML]{FFFFFF}} & 2.42\textbf{\color[HTML]{FFFFFF}} & 0.55\textbf{\color[HTML]{FFFFFF}} \\
    \cellcolor[HTML]{ECF4FF}+\textbf{Easy(ours)} & \cellcolor[HTML]{ECF4FF}28.3\textbf{\color[HTML]{4DAF4A}\scriptsize{(↓21.0)}} & \cellcolor[HTML]{ECF4FF}14.4\textbf{\color[HTML]{4DAF4A}\scriptsize{(↓9.5)}} & \cellcolor[HTML]{ECF4FF}4.0\textbf{\color[HTML]{4DAF4A}\scriptsize{(↓4.8)}} & \cellcolor[HTML]{ECF4FF}51.3\textbf{\color[HTML]{4DAF4A}\scriptsize{(↑1.0)}} & \cellcolor[HTML]{ECF4FF}21.1\textbf{\color[HTML]{4DAF4A}\scriptsize{(↓16.1)}} & \cellcolor[HTML]{ECF4FF}1.9\textbf{\color[HTML]{4DAF4A}\scriptsize{(↓2.4)}} & \cellcolor[HTML]{ECF4FF}66.9\textbf{\color[HTML]{4DAF4A}\scriptsize{(↑2.1)}} & \cellcolor[HTML]{ECF4FF}85.6\textbf{\color[HTML]{4DAF4A}\scriptsize{(↑2.1)}} & \cellcolor[HTML]{ECF4FF}\underline{89.2}\textbf{\color[HTML]{4DAF4A}\scriptsize{(↑2.6)}} & \cellcolor[HTML]{ECF4FF}2.52\textbf{\color[HTML]{4DAF4A}\scriptsize{(↑0.21)}} & \cellcolor[HTML]{ECF4FF}0.49\textbf{\color[HTML]{4DAF4A}\scriptsize{(↓0.06)}}\\
    \cellcolor[HTML]{ECF4FF}+\textbf{Medium(ours)} & \cellcolor[HTML]{ECF4FF}\underline{28.0}\textbf{\color[HTML]{4DAF4A}\scriptsize{(↓21.3)}} & \cellcolor[HTML]{ECF4FF}\underline{13.6}\textbf{\color[HTML]{4DAF4A}\scriptsize{(↓10.3)}} & \cellcolor[HTML]{ECF4FF}\underline{3.6}\textbf{\color[HTML]{4DAF4A}\scriptsize{(↓5.2)}} & \cellcolor[HTML]{ECF4FF}\underline{53.4}\textbf{\color[HTML]{4DAF4A}\scriptsize{(↑3.1)}} & \cellcolor[HTML]{ECF4FF}\underline{20.6}\textbf{\color[HTML]{4DAF4A}\scriptsize{(↓16.6)}} & \cellcolor[HTML]{ECF4FF}\textbf{1.2}\textbf{\color[HTML]{4DAF4A}\scriptsize{(↓3.1)}} & 
    \cellcolor[HTML]{ECF4FF}\underline{68.7}\textbf{\color[HTML]{4DAF4A}\scriptsize{(↑3.9)}} & \cellcolor[HTML]{ECF4FF}\textbf{87.3}\textbf{\color[HTML]{4DAF4A}\scriptsize{(↑3.8)}} & \cellcolor[HTML]{ECF4FF}\textbf{90.4}\textbf{\color[HTML]{4DAF4A}\scriptsize{(↑3.8)}} & \cellcolor[HTML]{ECF4FF}\textbf{2.84}\textbf{\color[HTML]{4DAF4A}\scriptsize{(↑0.53)}} & \cellcolor[HTML]{ECF4FF}\textbf{0.39}\textbf{\color[HTML]{4DAF4A}\scriptsize{(↓0.16)}}\\
    \cellcolor[HTML]{ECF4FF}+\textbf{Hard(ours)} & \cellcolor[HTML]{ECF4FF}\textbf{23.3}\textbf{\color[HTML]{4DAF4A}\scriptsize{(↓26.0)}} & \cellcolor[HTML]{ECF4FF}\textbf{11.5}\textbf{\color[HTML]{4DAF4A}\scriptsize{(↓12.4)}} & \cellcolor[HTML]{ECF4FF}{3.8}\textbf{\color[HTML]{4DAF4A}\scriptsize{(↓5.0)}} & \cellcolor[HTML]{ECF4FF}53.1\textbf{\color[HTML]{4DAF4A}\scriptsize{(↑2.8)}} & \cellcolor[HTML]{ECF4FF}{20.8}\textbf{\color[HTML]{4DAF4A}\scriptsize{(↓16.4)}} & \cellcolor[HTML]{ECF4FF}\underline{1.3}\textbf{\color[HTML]{4DAF4A}\scriptsize{(↓3.0)}} & \cellcolor[HTML]{ECF4FF}68.4\textbf{\color[HTML]{4DAF4A}\scriptsize{(↑3.6)}} & \cellcolor[HTML]{ECF4FF}\underline{87.2}\textbf{\color[HTML]{4DAF4A}\scriptsize{(↑3.7)}} & \cellcolor[HTML]{ECF4FF}\textbf{90.4}\textbf{\color[HTML]{4DAF4A}\scriptsize{(↑3.8)}} & \cellcolor[HTML]{ECF4FF}2.63\textbf{\color[HTML]{4DAF4A}\scriptsize{(↑0.32)}} & \cellcolor[HTML]{ECF4FF}\underline{0.46}\textbf{\color[HTML]{4DAF4A}\scriptsize{(↓0.09)}}\\
    \bottomrule
    \end{tabular}
    }
    \vspace{-4mm}
\end{table*}

Furthermore, our progressive curriculum design proves effective across all models, with performance steadily improving as training progresses through the Easy, Medium, and Hard stages. This confirms that the easy-to-hard learning trajectory offers a clear advantage over static data mixtures in enhancing overall performance.

\begin{wraptable}[22]{r}{0.48\columnwidth}
\vspace{-4mm}
\centering
\caption{Performance on general capability benchmarks. Best and second-best results are in \textbf{bold} and \underline{underlined}. Parentheses show absolute change from the corresponding baseline.}
\label{table2_main_result4}
\setlength{\tabcolsep}{4pt} 
\renewcommand{\arraystretch}{1.05}

\scriptsize 
\begin{tabular}{l|c|c}
\toprule
\textbf{Algorithm} & \textbf{LLaVA-wild} & \textbf{MMBench-CN} \\
\midrule

\multicolumn{1}{l|}{\textbf{LLaVA-v1.5-7B $^\S$}} & 55.7 & 0.646 \\

+HALVA $^\S$ & 61.0 & 0.589 \\

+DPO(SCPP) & 60.9 & 0.602 \\

\cellcolor[HTML]{ECF4FF}+\textbf{Easy(ours)} & \cellcolor[HTML]{ECF4FF}60.0{\scriptsize \textcolor[HTML]{4DAF4A}{(↑4.3)}} & \cellcolor[HTML]{ECF4FF}0.647{\scriptsize \textcolor[HTML]{4DAF4A}{(↑1e-3)}} \\

\cellcolor[HTML]{ECF4FF}+\textbf{Medium(ours)} & \cellcolor[HTML]{ECF4FF}\textbf{64.7}{\scriptsize \textcolor[HTML]{4DAF4A}{(↑9.0)}} & \cellcolor[HTML]{ECF4FF}\textbf{0.665}{\scriptsize \textcolor[HTML]{4DAF4A}{(↑1.9e-2)}} \\

\cellcolor[HTML]{ECF4FF}+\textbf{Hard(ours)} & \cellcolor[HTML]{ECF4FF}\underline{61.5}{\scriptsize \textcolor[HTML]{4DAF4A}{(↑5.8)}} & \cellcolor[HTML]{ECF4FF}\underline{0.664}{\scriptsize \textcolor[HTML]{4DAF4A}{(↑1.8e-2)}} \\

\midrule

\textbf{LLaVA-v1.6-7B $^\S$} & \underline{71.4} & 0.679 \\

+DPO(SCPP) & \underline{71.4} & 0.597 \\

\cellcolor[HTML]{ECF4FF}+\textbf{Easy(ours)} & \cellcolor[HTML]{ECF4FF}67.5{\scriptsize \textcolor[HTML]{E41A1C}{(↓3.9)}} & \cellcolor[HTML]{ECF4FF}0.671{\scriptsize \textcolor[HTML]{E41A1C}{(↓8e-3)}} \\

\cellcolor[HTML]{ECF4FF}+\textbf{Medium(ours)} & \cellcolor[HTML]{ECF4FF}70.8{\scriptsize \textcolor[HTML]{E41A1C}{(↓0.6)}} & \cellcolor[HTML]{ECF4FF}\textbf{0.682}{\scriptsize \textcolor[HTML]{4DAF4A}{(↑3e-3)}} \\

\cellcolor[HTML]{ECF4FF}+\textbf{Hard(ours)} & \cellcolor[HTML]{ECF4FF}\textbf{72.6}{\scriptsize \textcolor[HTML]{4DAF4A}{(↑1.2)}} & \cellcolor[HTML]{ECF4FF}\underline{0.681}{\scriptsize \textcolor[HTML]{4DAF4A}{(↑2e-3)}} \\

\midrule

\textbf{LLaVA-v1.5-13B $^\S$} & 64.9 & 0.685 \\

+HALVA $^\S$ & 63.3 & 0.636 \\

+DPO(SCPP) & 66.0 & 0.649 \\

\cellcolor[HTML]{ECF4FF}+\textbf{Easy(ours)} & \cellcolor[HTML]{ECF4FF}66.1{\scriptsize \textcolor[HTML]{4DAF4A}{(↑1.2)}} & \cellcolor[HTML]{ECF4FF}0.643{\scriptsize \textcolor[HTML]{E41A1C}{(↓4.2e-2)}} \\

\cellcolor[HTML]{ECF4FF}+\textbf{Medium(ours)} & \cellcolor[HTML]{ECF4FF}\textbf{68.6}{\scriptsize \textcolor[HTML]{4DAF4A}{(↑3.7)}} & \cellcolor[HTML]{ECF4FF}\textbf{0.701}{\scriptsize \textcolor[HTML]{4DAF4A}{(↑1.6e-2)}} \\

\cellcolor[HTML]{ECF4FF}+\textbf{Hard(ours)} & \cellcolor[HTML]{ECF4FF}\underline{67.4}{\scriptsize \textcolor[HTML]{4DAF4A}{(↑2.5)}} & \cellcolor[HTML]{ECF4FF}\underline{0.699}{\scriptsize \textcolor[HTML]{4DAF4A}{(↑1.4e-2)}} \\

\bottomrule
\end{tabular}

\end{wraptable}

\vspace{-4mm}

\paragraph{Evaluation on General Capabilities.}
A critical consideration for any alignment method is whether its gains in factuality come at the cost of general-purpose capabilities. We investigate this by evaluating our SCPO-tuned models on two diverse benchmarks: \textit{LLaVA-in-the-Wild} and \textit{MMBench-CN}. As reported in Table~\ref{table2_main_result4}, our results indicate that SCPO largely avoids the performance degradation observed in some baselines like HALVA, and in many cases, enhances the model's general abilities. For instance, on LLaVA-v1.5-7B, the model after the medium curriculum stage improves the \textit{LLaVA-wild score} from \textit{55.7} to \textit{64.7} and the \textit{MMBench-CN accuracy} from \textit{0.646} to \textit{0.665}. Similar positive trends are observed on the LLaVA-v1.5-13B model. While minor fluctuations exist across stages, the overall results strongly suggest that our method's focus on improving factual grounding leads to a more reliable and capable model in general, without inducing catastrophic forgetting.

\vspace{-3mm}
\subsection{Ablation Studies}
\vspace{-2mm}

\paragraph{Effectiveness of Objective Components.}
Our ablation of the SCPO objective in Table~\ref{tab:ablation_obj} demonstrates a clear synergistic effect between its components. While both the CCO and CSO objectives individually outperform the text-only baseline, their combination in the full SCPO objective is crucial for optimal performance. This unified approach achieves the best results across all metrics, reducing the \textit{CHAIRs} score by \textit{34.4\%} and the \textit{MMHal Bench HalRate} by \textit{39.2\%} relative to the text-only baseline. This confirms that the direct preference signals from CCO and the robust grounding enforced by CSO are complementary and essential for effective hallucination mitigation.

\begin{table*}[h]
\centering
\vspace{-3mm}
\setlength{\tabcolsep}{6pt}
\renewcommand{\arraystretch}{1.15}

\begin{minipage}[h]{0.48\textwidth}
\caption{Ablation on objective components. We evaluate the contribution of textual preference ($L_{\text{text}}$), image preference ($L_{\text{image}}$), and symmetric optimization ($L_{\text{CSO}}$). Best results are in \textbf{bold}.}
\label{tab:ablation_obj}
\vspace{-3mm}
\resizebox{\textwidth}{!}{
\begin{tabular}{ccc|cc|cc}
\toprule
\multicolumn{3}{c|}{\textbf{Objective Components}} 
& \multicolumn{2}{c|}{\textbf{Object Hallucination}} 
& \multicolumn{2}{c}{\textbf{MMHal Bench}} \\
\cmidrule(lr){1-3}\cmidrule(lr){4-5}\cmidrule(lr){6-7}
$L_{\text{text}}$ & $L_{\text{image}}$ & $L_{\text{CSO}}$
& \textbf{CHAIRs}\,$\downarrow$ & \textbf{CHAIRi}\,$\downarrow$
& \textbf{Score}\,$\uparrow$ & \textbf{HalRate}\,$\downarrow$ \\
\midrule
$\checkmark$ & $\checkmark$ & $\checkmark$ & \textbf{7.00} & \textbf{4.42} & \textbf{3.16} & \textbf{32.29} \\
$\checkmark$ & $\checkmark$ & $\times$     & 7.33  & 5.17  & 2.80 & 45.83 \\
$\checkmark$ & $\times$     & $\times$     & 10.67 & 5.91  & 2.55 & 53.13 \\
$\times$     & $\times$     & $\checkmark$ & 9.33  & 4.98  & 2.82 & 34.38 \\
\bottomrule
\end{tabular}
}
\end{minipage}\hfill
\begin{minipage}[h]{0.48\textwidth}
\caption{Ablation on training strategy components. \textbf{Sort}: data sorted by difficulty. \textbf{Update}: dynamic reference model update. Best results are in \textbf{bold}.}
\label{tab:ablation_strategy}
\vspace{-3mm}
\resizebox{\textwidth}{!}{
\begin{tabular}{cc|cc|cc}
\toprule
\multicolumn{2}{c|}{\textbf{Components}} 
& \multicolumn{2}{c|}{\textbf{Object Hallucination}} 
& \multicolumn{2}{c}{\textbf{MMHal Bench}} \\
\cmidrule(lr){1-2}\cmidrule(lr){3-4}\cmidrule(lr){5-6}
\textbf{Sort} & \textbf{Update}
& \textbf{CHAIRs}\,$\downarrow$ & \textbf{CHAIRi}\,$\downarrow$
& \textbf{Score}\,$\uparrow$ & \textbf{HalRate}\,$\downarrow$ \\
\midrule
$\checkmark$ & $\checkmark$ & \textbf{7.00} & \textbf{4.42} & 3.16 & \textbf{32.29} \\
$\checkmark$ & $\times$     & 9.67 & 5.88 & \textbf{3.22} & 37.50 \\
$\times$     & $\checkmark$ & 9.00 & 5.00 & 3.16 & 39.58 \\
$\times$     & $\times$     & 9.00 & 5.20 & 3.04 & 41.67 \\
\bottomrule
\end{tabular}
}
\end{minipage}

\vspace{-4mm}

\end{table*}

\paragraph{Effectiveness of the Curriculum Strategy.}
As shown in Table~\ref{tab:ablation_strategy}, our curriculum strategy's effectiveness stems from the synergy between data sorting (\textit{Sort}) and dynamic reference model updates (\textit{Update}). While each component provides moderate gains alone, their combination is critical. For example, adding sorting to dynamic updates further reduces the \textit{CHAIRs score} by \textit{22.2\%}. Similarly, adding dynamic updates to sorting lowers the \textit{MMHal Bench HalRate} by \textit{13.9\%}. This demonstrates that principled data scheduling and iterative model adaptation are mutually reinforcing, and both are essential to fully leverage our curriculum framework.

\vspace{-3mm}
\section{FURTHER ANALYSIS}
\vspace{-3mm}
\begin{wrapfigure}{r}{0.48\textwidth}
    \vspace{-15pt} 
    \centering
    
    \captionof{table}{Performance trajectory of the reversed curriculum. Best results are in \textbf{bold}.}
    \label{tab:reversed_curriculum}
    \resizebox{\linewidth}{!}{%
    \begin{tabular}{l|l|cc|cc}
    \toprule
    \multirow{2}{*}{\textbf{\makecell{Model \\ Params}}} & 
    \multirow{2}{*}{\textbf{\makecell{Training \\ Data}}} & 
    \multicolumn{2}{c|}{\textbf{Object HalBench}} & 
    \multicolumn{2}{c}{\textbf{MMHal Bench}} \\
    \cmidrule(lr){3-4} \cmidrule(lr){5-6}
    & & \textbf{CHAIRs}↓ & \textbf{CHAIRi}↓ & \textbf{Score}↑ & \textbf{HalRate}↓ \\
    \midrule
    $\theta_1$ & $\mathcal{D}_{\text{hard}}$   & \textbf{10.0} & \textbf{6.57} & 2.85 & \textbf{0.46} \\
    $\theta_2$ & $\mathcal{D}_{\text{medium}}$ & 11.3 & 6.61 & \textbf{2.93} & \textbf{0.46} \\
    $\theta_3$ & $\mathcal{D}_{\text{easy}}$   & 12.3 & 7.51 & 2.92 & 0.47 \\
    \bottomrule
    \end{tabular}%
    }

    \vspace{1.5em} 

    \includegraphics[width=\linewidth]{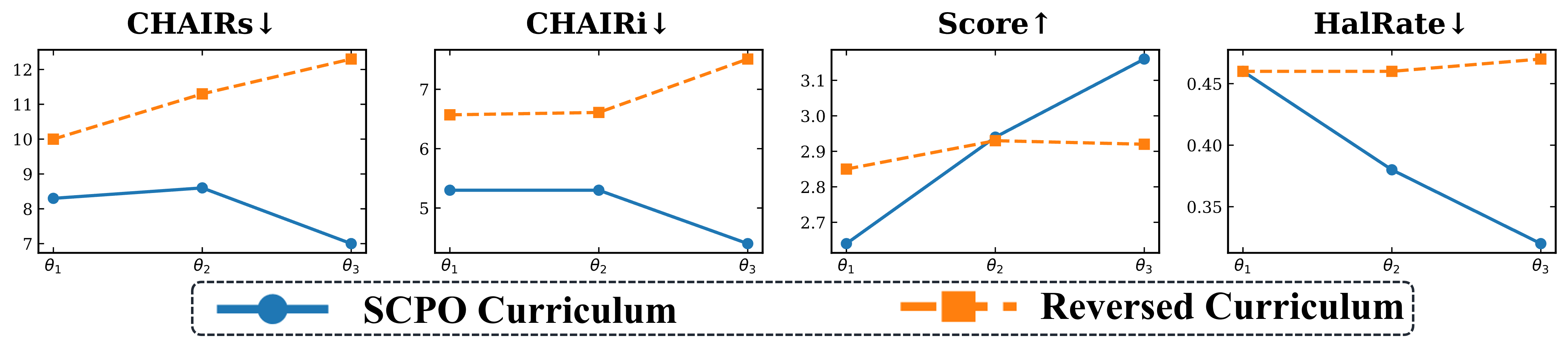}
    \vspace{-5mm} 
\captionof{figure}{Performance Trajectories of the Proposed and Reversed Curricula. The solid blue line represents our Easy-to-Hard curriculum, while the dashed orange line represents the reversed order. The x-axis denotes the curriculum stages, and the y-axis shows the performance on four key metrics}
    \label{fig:fig3}
\end{wrapfigure}
\paragraph{Effectiveness of the Curriculum Order.}
To verify the importance of our progressive curriculum, we evaluate a reversed Hard $\rightarrow$ Medium $\rightarrow$ Easy training sequence. The performance trajectory, detailed in Table~\ref{tab:reversed_curriculum} and visualized in Figure~\ref{fig:fig3}, reveals an unstable and suboptimal learning path. Notably, performance on \textit{Object HalBench} degrades as training progresses; the \textit{CHAIRs score} increases from an initial \textit{10.0} after the first stage to \textit{12.3} in the final model. This suggests that exposing the model to complex examples early, without a proper foundation, can impair its ability to handle fine-grained tasks. This unstable behavior is in sharp contrast to the steady improvements observed with our proposed Easy-to-Hard curriculum (shown in Table~\ref{table2_main_result3} and Figure~\ref{fig:fig3}). This finding confirms that the order of data difficulty is a critical factor in our framework, and highlights the need to build from simpler to more complex examples for robust learning.

\vspace{-3mm}
\section{Conclusion}
\vspace{-3mm}

In this work, we addressed the critical challenge of visual hallucinations in MLLMs by proposing SCPO, a novel alignment framework. SCPO's main contribution is a systematic approach that integrates three core innovations. First, we constructed the SCPP dataset, which provides semantically challenging preference pairs partitioned into a structured, easy-to-hard curriculum. Second, we designed a symmetric and bidirectional objective to enforce robust visual grounding. Finally, we introduced a curriculum-based training strategy where the model is trained progressively on data of increasing difficulty, coupled with a dynamic reference model that is updated at each stage. This unique combination of a data-driven curriculum and iterative alignment proves highly effective. Extensive experiments show that SCPO significantly reduces hallucinations across multiple benchmarks and model scales, establishing state-of-the-art performance while preserving general capabilities. Future work could explore applying this curriculum-based alignment paradigm to other modalities or different reasoning tasks.





Due to the double-blind review process, the \textbf{datasets and models} will be made publicly available after the conclusion of the review process. We will provide all necessary documentation to support reproducibility, including the data processing pipeline and model descriptions.

\bibliography{iclr2026_conference}
\bibliographystyle{iclr2026_conference}

\newpage
\appendix

\section{Algorithm Details}
\vspace{-3mm}
This section provides detailed pseudocode for our proposed SCPO framework. We first present the main training loop in Algorithm~\ref{alg:scpo_training}, which outlines the iterative, curriculum-based alignment process. Following that, we detail the procedure for computing the difficulty scores used to partition the dataset in Algorithm~\ref{alg:difficulty}.

\vspace{-2mm}
\begin{algorithm}[h]
\caption{Semantic Curriculum Preference Optimization (SCPO)}
\label{alg:scpo_training}
\begin{algorithmic}[1]
\Require Base model $\pi_{\theta_0}$, SCPP dataset $\mathcal{D}$, curriculum stages $S = \{\text{easy}, \text{medium}, \text{hard}\}$
\Require SCPO loss weights $\beta, \beta_1, \beta_2, \lambda$

\State Partition $\mathcal{D}$ into $\{\mathcal{D}_{\text{easy}}, \mathcal{D}_{\text{medium}}, \mathcal{D}_{\text{hard}}\}$ based on difficulty scores (Eq.~\ref{eq:difficulty_score})
\State Initialize policy model $\pi_{\theta} \gets \pi_{\theta_0}$
\State Initialize reference model $\pi_{\text{ref}} \gets \pi_{\theta_0}$

\For{stage in $S$}
    \State $\mathcal{D}_{\text{stage}} \gets \text{get\_data\_for\_stage}(\mathcal{D}, \text{stage})$
    \For{each batch $(I_w, I_l, y_w, y_l, Q) \in \mathcal{D}_{\text{stage}}$}
        \State // \textbf{Compute CCO Loss} (Eq.~\ref{eq:scpo_cco_optimization})
        \State $\mathcal{L}_{\text{text}} \gets -\log\sigma(\beta[\log\frac{\pi_\theta(y_w|I_w,Q)}{\pi_{\text{ref}}(y_w|I_w,Q)} - \log\frac{\pi_\theta(y_l|I_w,Q)}{\pi_{\text{ref}}(y_l|I_w,Q)}])$
        \State $\mathcal{L}_{\text{image}} \gets -\log\sigma(\beta[\log\frac{\pi_\theta(y_w|I_w,Q)}{\pi_{\text{ref}}(y_w|I_w,Q)} - \log\frac{\pi_\theta(y_w|I_l,Q)}{\pi_{\text{ref}}(y_w|I_l,Q)}])$
        \State $\mathcal{L}_{\text{CCO}} \gets \mathcal{L}_{\text{text}} + \mathcal{L}_{\text{image}}$
        
        \State // \textbf{Compute CSO Loss} (Eq.~\ref{eq:scpo_cso_optimization})
        \State $\mathcal{L}_{\text{match}}(I,y) \gets -\log\sigma(\beta_1[\log\frac{\pi_\theta(y|I,Q)}{\pi_{\text{ref}}(y|I,Q)} - \log\frac{\pi_\theta(y|Q)}{\pi_{\text{ref}}(y|Q)}])$
        \State $\mathcal{L}_{\text{contradict}}(I',y) \gets -\log\sigma(\beta_2[\log\frac{\pi_\theta(y|Q)}{\pi_{\text{ref}}(y|Q)} - \log\frac{\pi_\theta(y|I',Q)}{\pi_{\text{ref}}(y|I',Q)}])$
        \State $\mathcal{L}_{\text{CSO}} \gets \mathcal{L}_{\text{match}}(I_w,y_w) + \mathcal{L}_{\text{contradict}}(I_l,y_w) + \mathcal{L}_{\text{match}}(I_l,y_l) + \mathcal{L}_{\text{contradict}}(I_w,y_l)$
        
        \State // \textbf{Total SCPO Loss} (Eq.~\ref{eq:scpo_total_optimization})
        \State $\mathcal{L}_{\text{SCPO}} \gets \mathcal{L}_{\text{CCO}} + \lambda \mathcal{L}_{\text{CSO}}$
        \State Update $\theta$ using $\nabla_\theta \mathcal{L}_{\text{SCPO}}$
    \EndFor
    \State $\pi_{\text{ref}} \gets \pi_{\theta}$ \Comment{Dynamic reference model update}
\EndFor

\State \Return Final aligned model $\pi_{\theta}$
\end{algorithmic}
\end{algorithm}

\vspace{-3mm}
\begin{algorithm}[h]
\caption{Difficulty Score Computation for Preference Pairs}
\label{alg:difficulty}
\begin{algorithmic}[1]
\Require Preference pairs dataset $\mathcal{D} = \{(I_w^{(i)}, I_l^{(i)}, y_w^{(i)}, y_l^{(i)})\}_{i=1}^{N}$
\Require Weights $w_{\mathcal{H}}, w_{d_{\mathrm{OT}}}, w_{s_{\mathrm{CLIP}}} \in \mathbb{R}^+$
\Statex \textbf{Step 1: Feature Extraction}
\For{each pair $(I_w^{(i)}, I_l^{(i)})$ in $\mathcal{D}$}
    \State Compute MLLM mean entropy: $\overline{\mathcal{H}}^{(i)} \gets \frac{1}{2}(H(I_w^{(i)}) + H(I_l^{(i)}))$
    \State Compute DINOv2 optimal transport distance: $d_{\mathrm{OT}}^{(i)}$
    \State Calculate CLIP cosine similarity: $s_{\mathrm{CLIP}}^{(i)} \gets \langle \phi(I_w^{(i)}), \phi(I_l^{(i)}) \rangle$
\EndFor
\Statex
\Statex \textbf{Step 2: Standardization}
\For{each feature $f \in \{\overline{\mathcal{H}}, d_{\mathrm{OT}}, s_{\mathrm{CLIP}}\}$}
    \State Compute global statistics: $\mu_f, \sigma_f$
    \State Normalize: $z_f^{(i)} \gets (f^{(i)} - \mu_f) / \sigma_f$
\EndFor
\Statex
\Statex \textbf{Step 3: Weighted Aggregation}
\For{each pair $i=1, \dots, N$}
    \State $\text{Difficulty}^{(i)} \gets w_{\mathcal{H}} \cdot z_{\overline{\mathcal{H}}}^{(i)} + w_{d_{\mathrm{OT}}} \cdot z_{d_{\mathrm{OT}}}^{(i)} + w_{s_{\mathrm{CLIP}}} \cdot z_{s_{\mathrm{CLIP}}}^{(i)}$
\EndFor
\State \Return Difficulty scores $\{\text{Difficulty}^{(i)}\}_{i=1}^{N}$
\end{algorithmic}
\end{algorithm}

\section{SCPP Dataset Construction Details}
\label{app:data_details}

This section provides a detailed description of the data sources and the processing pipeline used to construct our \textit{Semantic Curriculum Preference Pairs (SCPP)} dataset, as mentioned in Section~\ref{sec:scpp_construction}.

\subsection{Data Sources and Rationale}
Our dataset is built by aggregating and refining four complementary public datasets. The selection of these sources is motivated by the need to create a diverse and challenging set of preference pairs that cover both textual and visual semantic contrasts.

\paragraph{Foundational Datasets.}
We start with two foundational datasets to provide a broad base of preference signals.
\begin{itemize}
    \item \textbf{RLHF-V}~\citep{yu2024rlhf}: This dataset offers large-scale prompt-text-image triples with fine-grained human feedback on textual hallucinations. It serves as our primary source for learning textual preferences, where the model must distinguish between a factually correct and a hallucinated description for a given image.
    \item \textbf{MVC}~\citep{wu2025symmetrical}: This dataset provides pairs of minimally contrastive images that differ in subtle but critical visual details (e.g., an object's color or count). It serves as our primary source for learning visual preferences, where the model must ground its response on the correct visual input.
\end{itemize}

\paragraph{Enhancement Datasets.}
Although the foundational datasets are useful, they have limitations. RLHF-V can contain noisy or low-quality images, and MVC's images are generated by diffusion models, which may lack the fidelity of real-world photographs. To address these issues and increase the dataset's quality and diversity, we incorporate two additional sources:
\begin{itemize}
    \item \textbf{HumanEdit}~\citep{bai2024humanedit}: This is a high-quality corpus of images edited by humans following specific instructions. It provides authentic and high-fidelity visual contrasts that are grounded in genuine human intent, serving as a source of realistic and challenging negative visual samples.
    \item \textbf{Winoground}~\citep{thrush2022winoground}: This benchmark is specifically designed to test visio-linguistic compositional reasoning. Its image-text pairs often differ by a single relational or compositional word (e.g., "the cat is behind the box" vs. "the box is behind the cat"). We include it to enrich SCPP with examples that require a deeper level of semantic understanding.
\end{itemize}

\subsection{LLM-based Text Generation for Preference Pairs}
\label{app:llm_pipeline}

While datasets like \textit{RLHF-V} and \textit{MVC} provide a strong foundation, other high-quality visual sources such as \textit{HumanEdit} and \textit{Winoground} lack pre-existing, preference-formatted textual annotations (i.e., a prompt with corresponding chosen and rejected responses). To take advantage of these valuable visual assets, we developed an LLM-based pipeline to generate high-quality, contextually appropriate text for their image pairs.

We utilized GPT-5 to perform this task. The pipeline was designed with three primary goals:
\begin{itemize}
    \item \textbf{Prompt Generation}: For each image pair, generate a concise and unambiguous question that targets the key semantic difference between them.
    \item \textbf{Response Generation}: Create a factually accurate "chosen" response that correctly describes the first image and a "rejected" response that correctly describes the second, ensuring both are brief and directly answer the generated question.
    \item \textbf{Quality Control}: Ensure the generated text is grammatically correct, stylistically consistent, and avoids ambiguous or subjective language. The process was carefully controlled to ensure all generated text is faithfully grounded in the visual content.
\end{itemize}

\begin{figure}[htp]
  \centering
  \includegraphics[width=\linewidth]{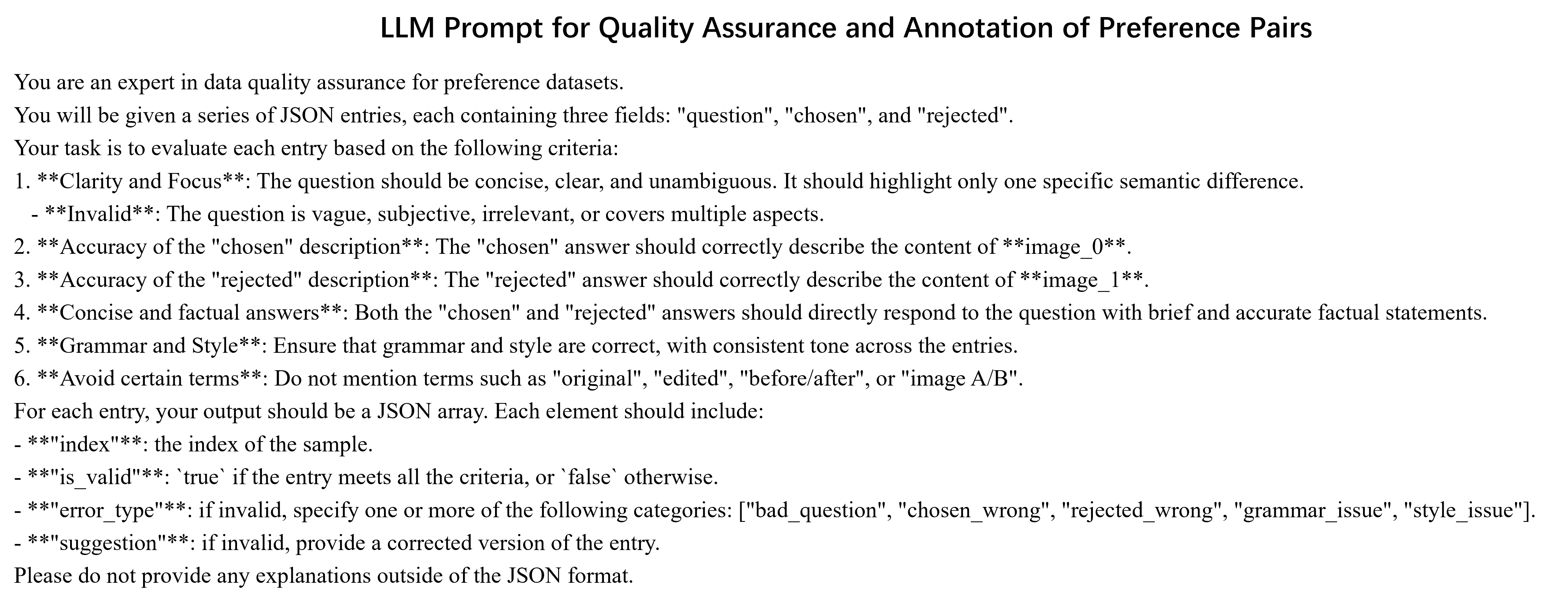}
  \caption{Data refinement prompt pipeline.}
  \label{fig:data_refinement_prompt}
\end{figure}

The core of this pipeline is a carefully designed prompt that guides the LLM to act as a data creator and quality assurance expert. The whole prompt set used in this process is provided in Figure~\ref{fig:data_refinement_prompt} for transparency and reproducibility.

\subsection{Difficulty Score Metrics}
\label{app:difficulty_metrics}

As introduced in Section~\ref{sec:scpp_construction}, our difficulty score is a composite measure that aggregates three complementary factors. Here, we provide their formal definitions and computational details for a given preference pair $(I_w, I_l)$.

\paragraph{MLLM Uncertainty ($\overline{\mathcal{H}}$).}
We quantify the model's uncertainty using the average binary semantic entropy over its forced-choice predictions for both the winning and losing images. A higher entropy value indicates greater ambiguity for the model. The average entropy is defined as follows:
\begin{equation}
    \overline{\mathcal{H}} = \tfrac{1}{2}\bigl(H(I_w) + H(I_l)\bigr),
\end{equation}
where $H(I)$ represents the binary entropy for a single image.

\paragraph{Semantic Proximity ($s_{\mathrm{CLIP}}$).}
This metric measures the high-level conceptual similarity between the two images. We compute it as the cosine similarity between their normalized feature embeddings, extracted from a pretrained CLIP vision encoder $\phi(\cdot)$~\citep{radford2021learning}:
\begin{equation}
    s_{\mathrm{CLIP}} = \langle \phi(I_w), \phi(I_l) \rangle.
\end{equation}
A higher score indicates that the images are strong semantic distractors.

\paragraph{Structural Discrepancy ($d_{\mathrm{OT}}$).}
To capture fine-grained, low-level visual differences, we measure the structural discrepancy. Let $\psi(I)$ denote the set of patch-level features extracted from an image $I$ using a DINOv2 encoder~\citep{oquab2024dinov2}. The discrepancy is then calculated as the Optimal Transport (OT) distance between the two sets of patch features:
\begin{equation}
    d_{\mathrm{OT}}(I_w, I_l) = \min_{T \in \Pi(\psi(I_w), \psi(I_l))} \sum_{i,j} T_{ij} C_{ij},
\end{equation}
where $C_{ij}$ is a joint feature-spatial cost matrix and $T$ is the optimal transport plan. A larger distance signifies a greater structural difference.

\paragraph{Final Score Computation.}
The final difficulty score is computed as the sum of the standardized values ($z$-scores) of the three metrics, as shown in Equation~\ref{eq:difficulty_score} in the main text. By standardizing each metric to have zero mean and unit variance, we place them on a comparable scale, allowing for a simple and unweighted aggregation. This approach avoids the need for additional hyperparameters for tuning the weights.

\subsection{Determining the Size of the Medium Curriculum Stage}
\label{app:data_ablation}

To determine the optimal size for the \textit{Medium} curriculum stage, we conducted an empirical study by varying the number of medium-difficulty samples used for training, ranging from \textit{1k} to \textit{10k}. The results, summarized in Table~\ref{tab:medium_size_ablation}, show a clear trend. As the data size increases from \textit{1k} to \textit{9k}, we observe a general improvement on most metrics, particularly in discriminative tasks. The model's performance on \textit{AMBER-Discriminative} peaks at the \textit{9k} data point, achieving the highest \textit{Accuracy} and \textit{F1 score}. While the lowest object hallucination (\textit{CHAIR}) is observed at the \textit{5k} mark, the \textit{9k} model maintains a highly competitive hallucination rate while offering superior discriminative capabilities. Notably, increasing the data size further to \textit{10k} leads to a slight degradation in several key metrics, including \textit{Cover} and \textit{Accuracy}, suggesting that \textit{9k} represents an optimal trade-off between performance, data diversity, and training efficiency. Based on this analysis, we set the size of the \textit{Medium} stage to \textit{9k} pairs for all main experiments.

\begin{table}[h]
\centering
\caption{Performance on AMBER with varying data sizes for the Medium curriculum stage. Best performance for each metric is highlighted in \textbf{bold}.}
\label{tab:medium_size_ablation}
\resizebox{0.8\textwidth}{!}{
\begin{tabular}{l|cccc|cc}
\toprule
\multicolumn{1}{c|}{\multirow{2}{*}{\textbf{Data Size}}} & \multicolumn{4}{c|}{\textbf{AMBER-Generative}} & \multicolumn{2}{c}{\textbf{AMBER-Discriminative}} \\
\cmidrule(lr){2-5} \cmidrule(lr){6-7}
~ & \textbf{CHAIR}↓ & \textbf{Cover}↑ & \textbf{HalRate}↓ & \textbf{Cog}↓ & \textbf{Acc.}↑ & \textbf{F1-Dis.}↑ \\ 
\midrule
1k  & 4.6 & 58.9 & 33.6 & 2.5 & 82.4 & 87.4 \\
2k  & 4.8 & 59.4 & 36.4 & 2.8 & 82.5 & 87.4 \\
3k  & 4.7 & 60.0 & 38.1 & 3.0 & 83.0 & 87.7 \\
5k  & \textbf{4.4} & 60.1 & 33.6 & 2.5 & 83.8 & 88.2 \\
7k  & 4.9 & \textbf{60.3} & 36.4 & 2.8 & 84.4 & 88.6 \\
9k  & 4.8 & \textbf{60.3} & \textbf{33.6} & \textbf{2.5} & \textbf{85.1} & \textbf{88.9} \\
10k & 4.6 & 58.7 & 38.1 & 3.0 & 84.8 & 88.7 \\
\bottomrule
\end{tabular}%
}
\end{table}

\section{Sensitivity Analysis of the CSO Weight $\lambda$}
\label{app:lambda_sensitivity}

The hyperparameter $\lambda$ in our SC-PO objective (Eq.~\ref{eq:scpo_total_optimization}) balances the contributions of the complementary ($\mathcal{L}_{\text{CCO}}$) and symmetric ($\mathcal{L}_{\text{CSO}}$) loss components. As these two objectives operate on different principles, their loss magnitudes can vary. The weight $\lambda$ is therefore crucial to prevent one component from dominating the optimization process.

To analyze the model's sensitivity to this parameter, we conducted a small-scale sweep for $\lambda$ on LLaVA-v1.6-7B. The results, summarized in Table~\ref{tab:lambda_ablation}, demonstrate that SCPO is remarkably robust to the choice of $\lambda$. Although $\lambda=0.2$ yields a only marginally better balance in terms of all metrics, its performance remains highly stable within the tested range of [0.1, 0.3]. For instance, the \textit{CHAIR} score only fluctuates between \textit{4.5} and \textit{4.6}. This low sensitivity is a significant practical advantage, indicating that the benefits of our framework are not contingent on meticulous hyperparameter tuning. Based on this robust performance, we selected $\lambda=0.2$ for all main experiments.

\begin{table}[h]
\centering
\caption{Sensitivity analysis for the CSO weight $\lambda$ on the AMBER benchmark. Performance remains highly stable, demonstrating the robustness of our method. Best performance is highlighted in \textbf{bold}.}
\label{tab:lambda_ablation}
\vspace{1mm}
\resizebox{0.75\textwidth}{!}{%
\begin{tabular}{c|ccc|cc}
\toprule
\multicolumn{1}{c|}{\multirow{2}{*}{\textbf{Setting ($\lambda$)}}} & 
\multicolumn{3}{c|}{\textbf{AMBER-Generative}} &  
\multicolumn{2}{c}{\textbf{AMBER-Discriminiative}} \\
\cmidrule(lr){2-4}\cmidrule(lr){5-6}
~ & \textbf{CHAIR}↓ & \textbf{Cover}↑ & \textbf{F1-Gen.}↑ & \textbf{Acc.}↑ & \textbf{F1-Dis.}↑ \\ \midrule
0.1 & 4.6 & 59.3 & 73.14 & 85.2 & 89.1  \\
\textbf{0.2} & \textbf{4.5} & \textbf{60.2} & \textbf{73.85} & \textbf{85.4} & \textbf{89.2} \\
0.3 & 4.6 & 60.0 & 73.67 & \textbf{85.4} & \textbf{89.2}  \\
\bottomrule
\end{tabular}%
}
\end{table}

\newpage
\section{Theoretical Analysis of Iterative Alignment}
\label{app:theoretical_analysis}

This appendix provides a concise motivation for our iterative alignment strategy with a \emph{dynamic reference} (Section~\ref{sec:iterative_learning}). We first analyze why a \emph{fixed} reference can fail in the late stages of curriculum, then derive the \emph{cumulative-reward} property of our approach.

\subsection{Limitations of a Fixed Reference Model in DPO}

A standard DPO setup with a fixed reference $\pi_{\text{ref}}$ can struggle when preference pairs in later curriculum stages are deliberately \emph{off-policy} for $\pi_{\text{ref}}$ (the preferred response has tiny probability under the reference).

\paragraph{Infinite KL (in practice, arbitrarily large).}
The objective regularized by the KL divergence  (cf.\ Eq.~\ref{eq:rlhf_objective}) requires a finite divergence between $\pi_\theta$ and $\pi_{\text{ref}}$. If $\pi_{\text{ref}}(y_w\mid x,I)\approx 0$ yet the optimized policy must assign nonzero mass to $y_w$, the divergence
\begin{equation}
D_{\mathrm{KL}}\!\big(\pi_\theta(\cdot\mid x,I)\,\Vert\,\pi_{\text{ref}}(\cdot\mid x,I)\big)
\end{equation}
can become arbitrarily large, rendering the optimization effectively intractable in practice.

\paragraph{Vanishing Gradients (Logistic Saturation).}
Let the per-pair DPO loss be $\ell(M)=-\log\sigma(\beta M)$ with margin
\begin{equation}
M_\pi(x,I)=
\log\frac{\pi(y_w\mid x,I)}{\pi(y_l\mid x,I)}
-
\log\frac{\pi_{\text{ref}}(y_w\mid x,I)}{\pi_{\text{ref}}(y_l\mid x,I)}.
\label{eq:dpo_margin}
\end{equation}
If the winner is severely off-policy under the fixed reference, i.e.,
$\pi_{\text{ref}}(y_w\mid x,I)\approx 0$ while $\pi_{\text{ref}}(y_l\mid x,I)$ is not, then
\begin{equation}
-\log\frac{\pi_{\text{ref}}(y_w\mid x,I)}{\pi_{\text{ref}}(y_l\mid x,I)}
\quad\text{is a large positive term.}
\label{eq:ref_log_ratio}
\end{equation}
However, if the new stage is initialized from the previous policy so that
$\pi \approx \pi_{\text{ref}}$, the two log-ratios approximately cancel and
\begin{equation}
M_\pi(x,I)\approx 0 \quad \text{at stage start, even when }\pi_{\text{ref}}(y_w\mid x,I)\text{ is tiny.}
\end{equation}
As training pushes the current log-ratio above the (tiny) reference log-ratio, $M_\pi(x,I)$ can grow large and positive, causing true saturation:
\begin{equation}
\frac{\partial \ell}{\partial M}
= -\beta\,\sigma(-\beta M)
\;\longrightarrow\; 0^{-}
\quad\text{as } M\to +\infty.
\label{eq:grad_pos_limit}
\end{equation}
For completeness, if instead $\pi_{\text{ref}}(y_l\mid x,I)\approx 0$ while $\pi_{\text{ref}}(y_w\mid x,I)$ is not, then $M_\pi(x,I)\to -\infty$ and
\begin{equation}
\frac{\partial \ell}{\partial M}\;\longrightarrow\;-\beta
\quad\text{as } M\to -\infty,
\label{eq:grad_neg_limit}
\end{equation}
i.e., the gradient does \emph{not} vanish. In late curriculum stages, the problematic regime is the former (winner off-policy under a fixed reference), where $M$ becomes very large and positive as $\pi$ departs from $\pi_{\text{ref}}$. Our dynamic reference update mitigates this by keeping $M$ in a moderate range at all stages.

\subsection{Cumulative Reward with a Dynamic Reference}
\label{sec:appendix_derivation}

We now show that resetting the reference at each stage, $\pi_{\text{ref}}^{(t+1)}\!\leftarrow\!\pi_t$, yields \emph{cumulative-reward} learning. Within stage $t$, we view the effective reward $r_t(x,I,y)$ as absorbing the weights from our CCO/CSO components. The DPO optimality condition implies the exponential family form:
\begin{equation}
\pi_t(y\mid x,I)\;\propto\;\pi_{\text{ref}}^{(t)}(y\mid x,I)\,
\exp\!\Big(\tfrac{1}{\beta}\,r_t(x,I,y)\Big).
\label{eq:stage_optimum}
\end{equation}

Let $\pi_0=\pi_{\text{Base}}$. Then:

\paragraph{Stage 1 (Easy).}
\begin{equation}
\pi_1 \;\propto\; \pi_0 \,\exp\!\Big(\tfrac{1}{\beta}\,r_1\Big).
\end{equation}

\paragraph{Stage 2 (Medium).}
\begin{equation}
\pi_2 \;\propto\; \pi_1 \,\exp\!\Big(\tfrac{1}{\beta}\,r_2\Big)
\;\propto\; \pi_0 \,\exp\!\Big(\tfrac{1}{\beta}(r_1+r_2)\Big).
\end{equation}

\paragraph{Stage 3 (Hard).}
\begin{equation}
\pi_3 \;\propto\; \pi_2 \,\exp\!\Big(\tfrac{1}{\beta}\,r_3\Big)
\;\propto\; \pi_0 \,\exp\!\Big(\tfrac{1}{\beta}(r_1+r_2+r_3)\Big).
\end{equation}

\paragraph{General $N$ stages.}
By induction,
\begin{equation}
\pi_N \;\propto\; \pi_0 \,\exp\!\Big(\tfrac{1}{\beta}\sum_{t=1}^N r_t\Big).
\label{eq:cumulative_reward}
\end{equation}
Thus, a single difficult optimization is decomposed into a sequence of tractable, on-policy stages whose rewards accumulate multiplicatively in the policy space. This also “resets” the KL at each stage boundary:
\begin{equation}
\mathbb{E}_{x,I}\,D_{\mathrm{KL}}\!\big(\pi_t(\cdot\mid x,I)\,\Vert\,\pi_{\text{ref}}^{(t+1)}(\cdot\mid x,I)\big)
=
\mathbb{E}_{x,I}\,D_{\mathrm{KL}}(\pi_t\Vert \pi_t)
=0,
\end{equation}
preventing cross-stage KL accumulation and helping avoid the saturation regime in Eq.~\eqref{eq:grad_pos_limit}.

\newpage
\section{Qualitative Examples}
\label{app:qualitative_examples}

To provide a more intuitive understanding of our SCPO framework's effectiveness, we present several qualitative examples drawn from our evaluation benchmarks. These examples illustrate how the model's responses evolve through the curriculum stages and how the final model ($\pi_{\text{Hard}}$) successfully mitigates hallucinations present in earlier stages or baseline models. In the following figures, text highlighted in \textcolor{red}{red} indicates a hallucination or factual error, while text in \textcolor{green}{green} indicates a correct, factually grounded statement.

\begin{figure}[h!]
    \centering
    \includegraphics[width=0.9\linewidth]{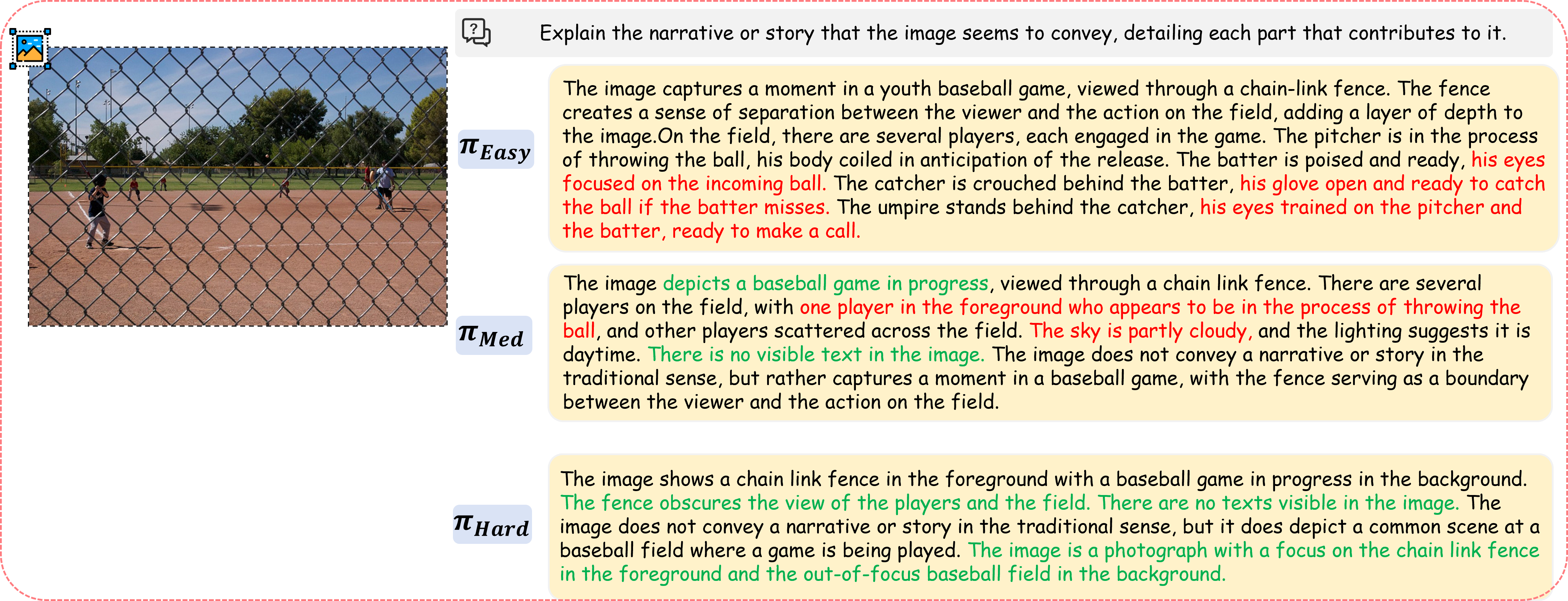}
    \caption{Example of mitigating descriptive hallucination. The initial model ($\pi_{\text{Easy}}$) hallucinates fine-grained details about the players' actions and focus. As the curriculum progresses, the model learns to be more faithful to the visual evidence, with the final model ($\pi_{\text{Hard}}$) correctly identifying that the fence obscures a clear view and avoids making unsubstantiated claims.}
    \label{fig:qual_baseball}
\end{figure}

\begin{figure}[h!]
    \centering
    \includegraphics[width=0.9\linewidth]{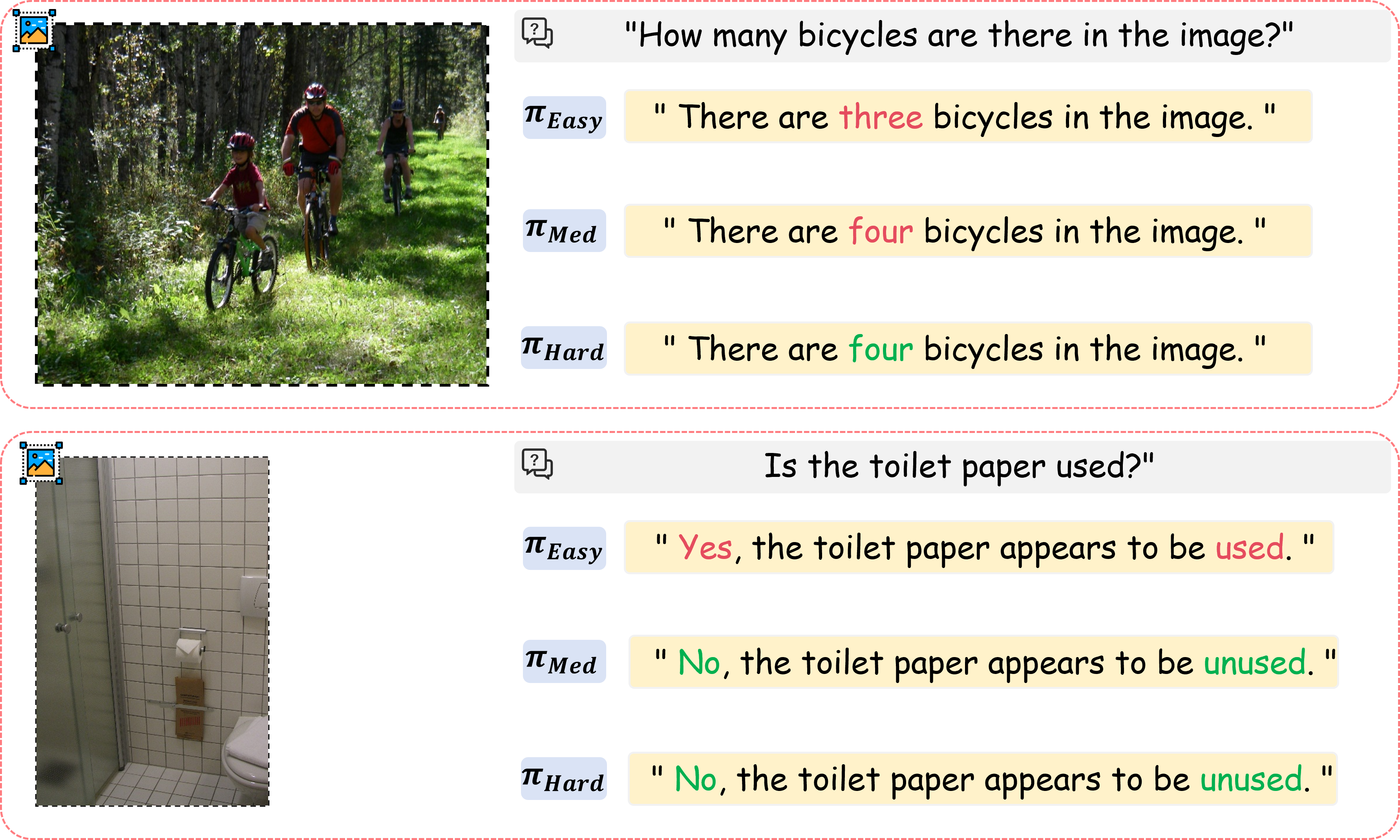}
    \caption{Examples of correcting factual errors. \textbf{(Top)} The model initially hallucinates the number of bicycles. Through curriculum learning, it corrects this counting error. \textbf{(Bottom)} The model initially makes an incorrect inference about the state of the toilet paper. The final model provides the correct, visually grounded answer.}
    \label{fig:qual_factual}
\end{figure}

\newpage

\section{Metric Computation Details}
\label{app:metric_details}

\paragraph{F1-Gen. Score for AMBER-Generative.}
Evaluating hallucination mitigation often involves a trade-off: a model might reduce hallucinations simply by generating shorter, less descriptive responses, which would unfairly penalize its \textit{Cover} score. The original AMBER benchmark provides separate metrics for hallucination (\textit{CHAIR}) and coverage (\textit{Cover}), but lacks a single score to evaluate this balance. To address this, we introduce the \textit{F1-Gen.} score, which provides a holistic measure of generative quality by harmonizing these two competing factors.

We frame the task of object generation in terms of precision and recall.
\begin{itemize}
    \item \textbf{Precision} reflects factuality. It is the fraction of generated objects that are correct, which we compute as $1 - \textit{CHAIR}$. High precision indicates low hallucination.
    \item \textbf{Recall} reflects comprehensiveness. It is the fraction of ground-truth objects mentioned in the response, which is directly measured by the \textit{Cover} metric.
\end{itemize}
The \textit{F1-Gen.} score is the harmonic mean of this precision and recall:
\begin{equation}
    \textit{F1-Gen.} = 2 \times \frac{(1 - \textit{CHAIR}) \times \textit{Cover}}{(1 - \textit{CHAIR}) + \textit{Cover}}.
\end{equation}
This composite metric allows for a fairer comparison between models, as it rewards methods that successfully reduce hallucinations (high precision) without sacrificing the descriptive richness of the output (high recall).

\section{Detailed Benchmark Results}
\label{app:detailed_results}

This section provides a more detailed breakdown of our model's performance on the AMBER and Object HalBench benchmarks, complementing the main results. These tables offer deeper insight into the consistent and thorough improvements brought by the SCPO framework.

\subsection{Detailed Performance on AMBER}
\label{app:amber_details}

Table~\ref{tab:appendix_amber_detailed} presents a detailed performance breakdown on the various semantic categories of the AMBER benchmark. The results show that our SCPO curriculum leads to a broad enhancement of fine-grained capabilities. Across nearly all model scales and sub-tasks (such as \textit{Action}, \textit{Attribute}, and \textit{Existence}), the models trained with our curriculum significantly outperform the original LLaVA baselines. The consistent improvement, particularly in the final stage, confirms our framework's ability to foster a more robust and nuanced semantic understanding.

\begin{table*}[h]
    \caption{Performance of different variants on the AMBER benchmark across model scales. Results are reported for each difficulty stage (Easy, Middle, Hard) and baseline models. Best and second-best results per metric and model group are \textbf{bolded} and \underline{underlined}, respectively.}
    \label{tab:appendix_amber_detailed}
    \vspace{-1mm}
    \resizebox{\textwidth}{!}{
    \begin{tabular}{l|c|c|c|c|c|c|c|c}
    \toprule
    \multicolumn{1}{c|}{\textbf{Model}} & \multicolumn{1}{c|}{\textbf{Discriminative}} & \multicolumn{1}{c|}{\textbf{Action}} & \multicolumn{1}{c|}{\textbf{Attribution}} & \multicolumn{1}{c|}{\textbf{Existence}} & \multicolumn{1}{c|}{\textbf{Number}} & \multicolumn{1}{c|}{\textbf{Relation}} & \multicolumn{1}{c|}{\textbf{State}} & \multicolumn{1}{c}{\textbf{Comprehensive\_Score}} \\
    \cmidrule(lr){1-9}

    LLaVA-v1.5-7B & 85.2 & 83.1 & 77.7 & 94.3 & \underline{83.4} & 69.1 & 74.0 & 74.94 \\
    +Easy & \underline{85.6} & \textbf{85.4} & \textbf{78.8} & \textbf{97.6} & \textbf{84.2} & 59.9 & {75.2} & 75.98 \\
    +Middle & \textbf{87.3} & \underline{84.5} & \underline{78.4} & \underline{97.2} & {82.2} & \underline{72.3} & \underline{76.0} & \underline{77.78} \\
    +Hard & \textbf{87.3} & 84.2 & 77.5 & 96.9 & 77.0 & \textbf{76.4} & \textbf{76.6} & \textbf{78.14} \\
    \midrule

    LLaVA-v1.6-vicuna-7B & 87.0 & \textbf{88.2} & \underline{81.5} & 95.1 & 83.0 & 69.6 & \underline{79.7} & 80.10 \\
    +Easy & 87.3 & \underline{86.9} & 80.9 & \textbf{97.8} & 83.2 & 64.5 & 79.0 & 80.25 \\
    +Middle & \underline{88.9} & \underline{86.9}& \textbf{82.6} & \underline{97.1} & \textbf{84.9} & \underline{73.0} & \textbf{80.9} & \underline{81.37} \\
    +Hard & \textbf{89.2} & {86.6} & \textbf{82.6} & 97.0 & \underline{84.8} & \textbf{75.3} & \textbf{80.9} & \textbf{81.52} \\
    \midrule

    LLaVA-v1.5-13B & 86.6 & 86.5 & 82.0 & 91.3 & 84.9 & \textbf{79.4} & 79.9 & 75.72 \\
    +Easy & \underline{89.2} & \textbf{89.3} & {84.7} & 95.7 & {89.5} & 73.8 & {81.8} & 78.03 \\
    +Middle & \textbf{90.4} & 88.1 & \underline{85.2} & \underline{96.6} & \underline{90.2} & \underline{77.9} & \underline{82.5} & \textbf{79.56} \\
    +Hard & \textbf{90.4} & \underline{88.2} & \textbf{85.4} & \textbf{97.0} & \textbf{90.9} & {76.4} & \textbf{82.6} & \underline{79.41} \\
    \bottomrule
    \end{tabular}
    }
    \vspace{-4mm}
\end{table*}

\subsection{Detailed Performance on Object HalBench}
\label{app:objhal_details}

Table~\ref{tab:appendix_objhal_detailed} provides a direct validation of primary goal of our framework to reduce hallucinations. The results show a clear and steady trend in all  models tested. As training progresses through the easy-to-hard curriculum, we observe a consistent and significant decrease in hallucination rates (both \textit{Response Hall.} and \textit{Object Hall.}), coupled with a corresponding increase in correctness scores. This provides clear evidence for the effectiveness of our progressive learning strategy in systematically improving the model factuality.

\begin{table*}[h]
    \caption{Detailed hallucination and correctness metrics on Object HalBench. Best and second-best results per metric and model group are in \textbf{bold} and \underline{underlined}.}
    \label{tab:appendix_objhal_detailed}
    \vspace{-1mm}
    \resizebox{\textwidth}{!}{
    \begin{tabular}{l|c|c|c|c}
    \toprule
    \multicolumn{1}{c|}{\textbf{Model}} & 
    \multicolumn{1}{c|}{\textbf{Response Hall. (↓)}} & 
    \multicolumn{1}{c|}{\textbf{Response Correct (↑)}} & 
    \multicolumn{1}{c|}{\textbf{Object Hall. (↓)}} & 
    \multicolumn{1}{c}{\textbf{Object Correct (↑)}} \\
    \cmidrule(lr){1-5}

    LLaVA-v1.5-7B & 55.97 & \underline{73.58} & 26.49 & 73.51 \\
    +Easy & {29.12} & {70.88} & {14.98} & {85.02} \\
    +Middle &  \underline{26.48} & {73.52} & \underline{13.33} & \underline{86.67} \\
    +Hard & \textbf{21.40} & \textbf{78.60} & \textbf{11.65} & \textbf{88.35} \\
    \midrule

    LLaVA-v1.6-vicuna-7B & 12.95 & 87.05 & 6.79 & 93.21 \\
    +Easy & \underline{9.03} & \underline{90.97} & {5.31} & {94.69} \\
    +Middle & {9.25} & {90.75} & \underline{5.28} & \underline{94.72} \\
    +Hard & \textbf{7.61} & \textbf{92.39} & \textbf{4.42} & \textbf{95.58} \\
    \midrule

    LLaVA-v1.5-13B & 50.86 & 49.14 & 23.86 & 76.14 \\
    +Easy & {29.41} & {70.59} & {14.44} & {85.56} \\
    +Middle &  \underline{28.97} & \underline{71.03} & \underline{13.58} & \underline{86.42} \\
    +Hard & \textbf{24.56} & \textbf{75.44} & \textbf{11.51} & \textbf{88.49} \\
    \bottomrule
    \end{tabular}
    }
    \vspace{-4mm}
\end{table*}

\section{LLM Usage}

In this work, we used Large Language Models (LLMs) solely for language refinement to improve the clarity and fluency of the text. The LLM played no role in the ideation, analysis, or generation of scientific content.

\end{document}